\documentclass[11pt]{article}
\usepackage{graphicx}
\usepackage{amssymb}
\usepackage{amscd}
\usepackage{mathrsfs}
\usepackage{longtable,lscape}
\usepackage{amsthm}
\usepackage{amsfonts}
\usepackage{amsmath}
\usepackage{bbm}
\usepackage{float}
\usepackage{url}
\usepackage[round]{natbib}

\newcommand{\compl}{{\mathbb C}}

\newcommand{\captionfonts}{\footnotesize}
\makeatletter 
\long\def\@makecaption#1#2{%
\vskip\abovecaptionskip
\sbox\@tempboxa{{\captionfonts #1: #2}}%
\ifdim \wd\@tempboxa >\hsize
{\captionfonts #1: #2\par}
\else
\hbox to\hsize{\hfil\box\@tempboxa\hfil}%
\fi
\vskip\belowcaptionskip}
\makeatother 
\begin{document}
\title{Beyond-Quantum Modeling of Question Order Effects and Response Replicability in Psychological Measurements}
\author{Diederik Aerts$^1$ and Massimiliano Sassoli de Bianchi$^{2}$ \vspace{0.5 cm} \\ 
\normalsize\itshape
$^1$ Center Leo Apostel for Interdisciplinary Studies \\
\normalsize\itshape
Brussels Free University, 1050 Brussels, Belgium \\ 
\normalsize
E-Mail: \url{diraerts@vub.ac.be}
\\ 
\normalsize\itshape
$^2$ Laboratorio di Autoricerca di Base \\ 
\normalsize\itshape
6914 Lugano, Switzerland \\
\normalsize
E-Mail: \url{autoricerca@gmail.com} \\
}
\date{}
\maketitle
\begin{abstract}
\noindent 
A general tension-reduction (GTR) model was recently considered \citep{AertsSassolideBianchi2015a,AertsSassolideBianchi2015b} to derive quantum probabilities as (universal) averages over all possible forms of non-uniform fluctuations, and explain their considerable success in describing experimental situations also outside of the domain of physics, for instance in the ambit of quantum models of cognition and decision. Yet, this result also highlighted the possibility of observing violations of the predictions of the Born rule, in those situations where the averaging would not be large enough, or would be altered because of the combination of multiple measurements. In this article we show that his is indeed the case in typical psychological measurements exhibiting question order effects, by showing that their statistics of outcomes are inherently non-Hilbertian, and require the larger framework of the GTR-model to receive an exact mathematical description. We also consider another unsolved problem of quantum cognition: response replicability. It is has been observed that when question order effects and response replicability occur together, the situation cannot be handled anymore by quantum theory. However, we show that it can be easily and naturally described in the GTR-model. Based on these findings, we motivate the adoption in cognitive science of a hidden-measurements interpretation of the quantum formalism, and of its GTR-model generalization, as the natural interpretational framework explaining the data of psychological measurements on conceptual entities. 
\end{abstract}
\\

\section{Introduction}

Quoting from Wikipedia:\footnote{Quantum cognition (n.d.). In Wikipedia. Retrieved August 10, 2015, from \url{https://en.wikipedia.org/wiki/Quantum_cognition}.} ``Quantum cognition is an emerging field which applies the mathematical formalism of quantum theory to model cognitive phenomena such as information processing by the human brain, decision making, human memory, concepts and conceptual reasoning, human judgment, and perception.'' \citep{Kitto2008, Khrennikov2010, BusemeyerBruza2012, AertsEtal2013a, AertsEtal2013b, PothosBusemeyer2013, WangEtal2013,BlutnerbeimGraben2014}. 

To explain why quantum physics is so adequate in the modeling of the human cognitive processes, many reasons are today available that are beginning to be fairly well understood. An important one is that quantum mechanics was created to model and explain physical phenomena exhibiting a high degree of contextuality, and contextuality is also the light motive in human cognition and decision processes. For instance, in the same way quantum superposition states can describe physical properties only existing in potential terms, that is, available to be actualized during certain measurements in a genuinely unpredictable way, conceptual situations that lend themselves to judgments and decisions are also generally describable in terms of indeterministic processes of actualization of potential properties, when subjects are somehow forced to produce an answer in situations of ambiguity, uncertainty, confusion, etc. 

So, the quantum formalism can be very relevant in the description of the human cognitive processes, and it was quite amazing to observe, over the last years, how many aspects that were used to explain the strange behavior of the microscopic physical entities have proven to be also pertinent in the description of the human conceptual entities, like superposition, interference, entanglement, emergence, and even quantum fields effects. The amazement was not only in the observation that the human conceptual entities and the microscopic physical entities were evidently sharing a common ``conceptual nature,''\footnote{This observation also brought one of us to boldly propose a new \emph{conceptuality interpretation} of quantum mechanics \citep{Aerts2009,Aerts2010a,Aerts2010b}.} but that the very specific Hilbertian structure of quantum mechanics appeared to be sufficient to model a large spectrum of experimental situations.

In other terms, it wasn't at all expected in the beginning of the investigations in quantum cognition, that the Born rule of assignment of the probabilities would prove to be so effective in the modeling of many quantum effects identified in the cognitive domain \citep{AertsSozzo2012a, AertsSozzo2012b}. To put it another way: Why the Born rule and not other ``rules,'' associated with different and/or more general quantum-like structures? This fundamental question was the object of a recent examination by us: starting from a beyond-quantum modeling of psychological measurements, which we have called the \emph{general tension-reduction model} (GTR-model, in brief), we were able to show that the Born rule is characterizable in terms of uniform fluctuations in the experimental context, and that these uniform fluctuations naturally emerge when a \emph{universal average} over all possible forms of non-uniform fluctuations is considered \citep{AertsSassolideBianchi2015a,AertsSassolideBianchi2015b}. In other terms, in our analysis we were able to show that Born's quantum probabilities correspond to a first-order approximation of more general probability models, describing a beyond-quantum class of measurements characterized by non-uniform fluctuations.

Let us explain a little more specifically what we exactly mean by this, as this will be important to properly contextualize the results presented in this article, which can be considered to be the logical continuation of the analysis in  \citet{AertsSassolideBianchi2015a,AertsSassolideBianchi2015b}. Consider an opinion poll conducted on a large number of respondents. Each of them will bring into it the uniqueness of their minds, each one with a specific conceptual network forming its inner memory structure. This means that participants in the poll will have a different way of choosing an answer among those that are available to them to be selected. And this also means that a cognitive experiment, involving a number of different subjects, is generally to be considered not as a single measurement, but as a collection of different measurements, one for each participant. Each of these different individual measurements, however, is in principle associated with different outcome probabilities, so that the overall probabilities deduced from the individual results produced by all the participants are in fact averages over the individual probabilities. 

What we are here affirming is that in a typical cognitive experiment, like an opinion poll, we are actually in the presence of a sort of abstract ``collective mind,'' producing a collection of different quantum-like measurements\footnote{We use the term ``quantum-like'' in the sense of ``non-classical \& non-quantum,'' i.e., to describe processes that can be described by probability models which are  simultaneously non-Kolmogorovian and non-Hilbertian.} that remain undistinguished, being simply averaged out in the final statistics. This means that we are in a situation of lack knowledge about the different individual `ways of choosing' employed by the individual respondents, each one characterized by a different way of assigning probabilities to the available outcomes. Now, considering the general success of the Born rule in the modeling of data gathered from experiments which are, as we said, statistical mixtures of different measurements, one may expect to be able to understand the quantum mechanical Born rule as a rule describing the probabilistics of outcomes of measurements that are mixtures of a sufficiently vast ensemble of different types of measurements. 

This is precisely the result that we have recently obtained \citep{AertsSassolideBianchi2015a,AertsSassolideBianchi2015b}: starting from the general framework provided by the GTR-model, we were able to define, in a mathematically precise and cognitively transparent way, a notion of \emph{universal measurement}, describing the most general possible condition of lack of knowledge in a measurement, expressed as an average over all imaginable types of individual measurements. We have then proved that such huge average exactly yields the Born rule of quantum mechanics, if the structure of the state space is Hilbertian, providing in this way a convincing explanation of why the quantum statistics performs so well, in so many experimental ambits. It does so because it is a first order non-classical theory in the modeling of measurement data. 

However, if on one hand the possibility to describe the Born rule as a universal measurement can explain its ``unreasonable'' success in modeling so many experiments in cognitive science, this also points to the possibility of observing violations of it, in specific experimental situations. Indeed, there are no \emph{a priori} reasons to believe that in all experiments a universal average, or a good approximation of it, would be generated by the participants' different `ways of choosing' an outcome. This can be the case because the statistical sample is too small, because at the inter-subjective level certain `meaning connections' would be absent, thus filtering out certain ``ways of choosing,'' or because measurements would be combined together in a way that would prevent the average to be effective in producing the uniform fluctuations that characterize the Born rule. 

It is important to observe that possible violations of the Born rule cannot be observed in single measurement situations, as is clear that the Hilbert model, with its scalar product, when equipped with the Born rule becomes a ``universal probabilistic machine,'' capable of representing all possible probabilities appearing in nature, in a given (single) measurement context \citep{AertsSozzo2012a,AertsSassolideBianchi2015a}. But when for instance
sequential measurements are considered, the Hilbertian model is not any more a universal model, and violations can in principle be observed \citep{AertsSassolideBianchi2015a,AertsSassolideBianchi2015b}. So, it is in the ambit of experiments constructed as  multiple processes, where aspects of non-commutativity are at stake, that one has to look to find quantum-like structures which are possibly non-Hilbertian, i.e., non-Bornian.

Typical experiments of this kind are those exhibiting \emph{question order effects}, commonly observed in social and behavioral research; see for instance 
\citet{SudmanBradburn1974, SchumanPresser1981, TourangeauRipsRasinski2000}. Tentative quantum accounts of these and other effects were given by many authors in recent years \citep{ConteEtal2009, BusemeyerEtal2011, TruebloodBusemeyer2011, AtmanspacherRoemer2012, WangBusemeyer2013, AertsEtal2013a, AertsEtal2013b, WangEtal2014, KhrennikovEtal2014, Boyer-KassemEtal2015} and it is generally considered that quantum mechanics can provide an efficacious and consistent modeling of question order effects data. This was recently emphasized by \citet{WangBusemeyer2013}, showing that different data sets, like those reported by \citet{Moore2002}, obey a parameter-free constraint, called the \emph{QQ-equality}, which according to these authors is able to test the pure quantum nature of the experimental probabilities. 

This conclusion, however, was recently called into question by \citet{Boyer-KassemEtal2015}, showing that the QQ-equality is actually insufficient to test the validity of the Hilbertian model, as other `quantum equalities' can be shown not to be satisfied by the same data, thus demonstrating a violation of the Hilbertian structure. But then, how to properly account for the question order effects, if the Hilbertian model is unable to do so? Because one thing is to show that the Born rule is violated, and another thing is to provide a \emph{non ad hoc} model able to exactly fit the experimental data and, possibly, to shed some light into the nature of the cognitive processes that have generated the data. 

This is precisely what we are going to do in the present article, by means of the GTR-model. More precisely, in Section~\ref{2-outcome-measurements}, we show how two sequential $2$-outcome measurements can be generally modeled by using the hidden-measurements representation that is 
built in the GTR-model. In Section~\ref{simplifying hypotheses}, we add more structure to the model by introducing three simplifying hypothesis (that we call weak compatibility, local uniformity and sensitivity to pre-measurement state) thanks to which, in Section~\ref{solving the model}, we can present an explicit exact solution to the model. This solution depends on a certain number of parameters (only one of which is free), and standard quantum mechanics is recovered in the limit where some of these parameters tend to certain specific values.

In Section~\ref{modeling the data}, we use this explicit solution to obtain an exact modeling of some typical data from the Gallup survey experiments reported in a seminal article on question order effects by \cite{Moore2002}, and more particularly the `Clinton/Gore' and `Rose/Jackson' data. Thanks to our exact modeling, we can then show that the experimental probabilities are irreducibly non-Hilbertian, as they cannot be obtained by a choice of parameters defining a pure quantum situation. In Section~\ref{quantum tests}, we also investigate the content of Wang \& Busemeyer’s QQ-equality, showing how and why it can also be obeyed by non-Hilbertian models. We also use our explicit solution to derive additional parameter-free `quantum equalities,' which we show are strongly violated by the Clinton/Gore experimental data. 

In Section~\ref{response replicability}, we address the issue of \emph{response replicability}, which remains an unsolved problem in quantum cognition. Indeed, the quantum formalism does not allow the joint description of question order effects and response replicability \citep{KhrennikovEtal2014}. This is so because the quantum model, contrary to the GTR-model, does not allow for the description of different typologies of measurements, associated with a same set of outcomes. On the other hand, by allowing the measurements to ``evolve,'' taking into account the obtained outcomes, the GTR-model can 
not only describe response replicability, compatibly with question order effects, but also more complex situations of `partial replicability,' in accordance with our intuitive idea of what it means to have an opinion, or if we had not to have formed one, and to possibly change it again, under the right circumstances. 

In Section~\ref{interpretational framework}, we then provide what we think is the appropriate interpretation of the different mathematical objects  of
the GTR-model,  with the states not describing beliefs, but objective elements of a conceptual reality, independent from the minds that can interact with it. 
And considering that the GTR-model is a natural generalization of the standard quantum formalism, we encourage the adoption of this interpretation as the standard one in quantum cognition. In Section~\ref {Averaging}, we briefly consider what are the possible effects of the averaging over the different respondents, for what concerns the effective probability model describing the overall statistics of outcomes. In particular, we show that even when the individual respondents choose their outcomes according not only to a symmetrical probability distribution, but also in the same way in the two sequential measurements, when their `ways of choosing' are averaged out, in the final statistics, this symmetry will be broken, in the sense that their collective `way of choosing' will be non-symmetric and sequentially non-uniform.  Finally, in Section~\ref{concluding remarks}, we recap the obtained results, and offer some final thoughts regarding the possibility of modeling question order effects by means of the Born rule, in higher dimensional Hilbert spaces.

\section{Sequential $2$-outcome measurements}
\label{2-outcome-measurements}

Let us start by considering two general measurements, $F$ and $G$, performed on an entity $S$, prepared in the initial state $\psi$. We assume that each measurement admits 4 different possible outcomes: $F1$, $F2$, $F3$, $F4$, and $G1$, $G2$, $G3$, $G4$, respectively. We denote $P(Fi|\psi)$ the probability of obtaining outcome $Fi$, $i=1,\dots, 4$, when measurement $F$ is performed, conditional to the fact that the pre-measurement state is $\psi$, and similarly, we denote $P(Gi|\psi)$ the probability of obtaining outcome $Gi$, $i=1,\dots, 4$, when measurement $G$ is performed, with the entity in the initial state $\psi$. Since for each measurement the 4 outcomes describe mutually exclusive events, which are also assumed to be the only possible events, then according to the properties of unitarity and additivity of a probability measure, we have: 
\begin{equation}
\sum_{i=1}^4 P(Fi|\psi) =1,\quad\quad \sum_{i=1}^4 P(Gi|\psi) =1.
\label{unitarity}
\end{equation}
Apart from the above two constraints, and the fact that probabilities must be positive real numbers, no other constraints need to be imposed on the specific values taken by the above eight probabilities, considering that $F$ and $G$ are two different arbitrary measurements.

Let us now assume that $F$ and $G$ are two composite measurements, corresponding to the sequential execution of two different measurements, $A$ and $B$, each one admitting two possible outcomes: $Ay$, $An$, and $By$, $Bn$, respectively (a $2$-outcome measurement is sometimes called a ``yes-no'' measurement, hence the letters ``y'' and ``n'' to designate the outcomes). More precisely, $F=AB$ is the measurement obtained by first performing measurement $A$ and then, on the resulting outcome state, measurement $B$ (note that the notation ``$F=AB$'' is here just a mnemonic device and is not meant to designate any specific operatorial product). Therefore, the 4 different possible outcomes of $F$ are: $F1=AyBy$, $F2=AyBn$, $F3=AnBy$ and $F4=AnBn$. On the other hand, $G=BA$ is the measurement obtained by first performing the measurement $B$ and then, on the resulting outcome state, measurement $A$. Therefore, the four possible outcomes are: $G1=ByAy$, $G2=ByAn$, $G3=BnAy$ and $G4=BnAn$. The unitarity relations (\ref{unitarity}) then read: 
\begin{eqnarray}
&&P(AyBy|\psi) + P(AyBn|\psi) + P(AnBy|\psi) + P(AnBn|\psi) =1,\nonumber\\
&&P(ByAy|\psi) + P(ByAn|\psi) + P(BnAy|\psi) + P(BnAn|\psi) =1.
\label{unitarity2}
\end{eqnarray}

We can write the above $8$ probabilities in more explicit terms by introducing the conditional probabilities relative to the second measurement. More precisely, denoting $P(By|\psi Ay)$ the probability that outcome $By$ is obtained in the second $B$-measurement, when outcome $Ay$ was obtained in the first $A$-measurement, which in turn was performed on the initial state $\psi$ (hence the sequential notation ``$\psi Ay$'' for the conditional statement, to be read from the left to the right), and so on, we can write: 
\begin{eqnarray}
&&P(AyBy|\psi)=P(By|\psi Ay) P(Ay|\psi), \quad P(AyBn|\psi) =P(Bn|\psi Ay)P(Ay|\psi),\nonumber\\
&&P(AnBy|\psi) = P(By|\psi An)P(An|\psi),\quad P(AnBn|\psi) = P(Bn|\psi An)P(An|\psi).
\label{conditionalAB}
\end{eqnarray}
In the same way, the probabilities for the outcomes of the $G=BA$ sequential measurement can be written as: 
\begin{eqnarray}
&&P(ByAy|\psi)= P(Ay|\psi By)P(By|\psi), \quad P(ByAn|\psi) =P(An|\psi By)P(By|\psi),\nonumber\\
&&P(BnAy|\psi) = P(Ay|\psi Bn)P(Bn|\psi),\quad P(BnAn|\psi) = P(An|\psi Bn)P(Bn|\psi),
\label{conditionalBA}
\end{eqnarray}
and we now have the four unitarity relations: 
\begin{eqnarray}
&&P(By|\psi Ay)+P(Bn|\psi Ay) = 1,\quad P(By|\psi An)+P(Bn|\psi An)=1,\nonumber\\
&&P(Ay|\psi By)+P(An|\psi By)=1,\quad P(Ay|\psi Bn) + P(An|\psi Bn)=1.
\label{unitarity3}
\end{eqnarray}

Clearly, the above remains a very general representation, not introducing any specific constraint on the structure of the outcome probabilities of $F$ and $G$. In other terms, until something more specific is said about the nature of the measured entity and the characteristics of the two measurements $A$ and $B$, it is always possible to describe two arbitrary $4$-outcome measurements, $F$ and $G$, as two sequential measurements formed by two $2$-outcome measurements, performed in different order, i.e. as: $F=AB$ and $G=BA$.

\subsection{The hidden-measurements representation}

We now introduce a very general geometrico-dynamical representation of the two sequential measurements $F=AB$ and $G=BA$. We emphasize from the beginning, to avoid possible misunderstandings, that this is a general quantum-like representation, generalizing both the classical and quantum probability models, which can be recovered in the appropriate limits.

It is worth observing that variants of this geometrico-dynamical representation  have received different names, in different research contexts. When describing $2$-outcome processes, as it will be the case in the present article, it was named  the \emph{sphere-model} \citep{Aertsetal1997} and also  the \emph{$\epsilon$-model} \citep{Aerts1998, SassolideBianchi2013}. When describing more general $N$-outcome measurements with uniform fluctuations, it was called the \emph{quantum model theory} \citep{AertsSozzo2012a, AertsSozzo2012b}, and when more general non-uniform fluctuations were also considered, in a recent update, also including degenerate measurements, it was named  the \emph{general tension-reduction model} \citep{AertsSassolideBianchi2015a,AertsSassolideBianchi2015b}. Also, when the structure of the state space is purely Hilbertian, as it is the case in microphysics, the model was called the \emph{extended Bloch representation of quantum mechanics} \citep{AertsSassolideBianchi2014, AertsSassolideBianchi2015c}, and offers a challenging solution to the measurement problem for finite-dimensional quantum systems of arbitrary dimension.

\begin{figure}[!ht]
\centering
\includegraphics[scale =.3]{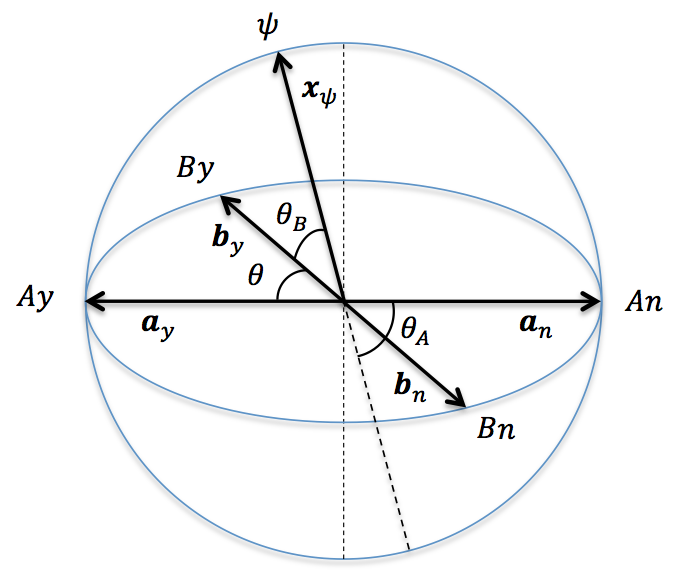}
\caption{The unit vectors ${\bf a}_y$, ${\bf a}_n=-{\bf a}_y$, representing the outcomes $Ay$ and $An$ of the $A$-measurement, the unit vectors ${\bf b}_y$, ${\bf b}_n=-{\bf b}_y$, representing the outcomes $By$ and $Bn$ of the $B$-measurement, and the unit vector ${\bf x}_\psi$, representing the initial state $\psi$, in the Bloch sphere. The angles $\theta$, $\theta_A$ and $\theta_B$ are given by the scalar products: ${\bf x}_\psi\cdot {\bf a}_y = \cos\theta_A$, ${\bf x}_\psi\cdot {\bf b}_y = \cos\theta_B$, and ${\bf a}_y\cdot {\bf b}_y = \cos\theta$.
\label{Bloch}}
\end{figure}

We mathematically describe the pre-measurement state $\psi$ of the entity subjected to the measurements 
by a $3$-dimensional real unit vector ${\bf x}_\psi$, i.e., by a vector at the surface of a $3$-dimensional unit sphere, called the \emph{Bloch sphere}. 
Measurements will then be represented in the sphere by abstract $1$-dimensional breakable and elastic structures, anchored at two antipodal points, corresponding to the two possible outcomes. More precisely, measurement $A$ will be represented by a breakable ``elastic band'' stretched between the two points ${\bf a}_y$ and ${\bf a}_n = -{\bf a}_y$, $\|{\bf a}_y\| = \|{\bf a}_n\|=1$, corresponding to the two outcomes $Ay$ and $An$, respectively. Similarly, measurement $B$ will be represented by a breakable ``elastic band'' stretched between the two points ${\bf b}_y$ and ${\bf b}_n = -{\bf b}_y$, $\|{\bf b}_y\| = \|{\bf b}_n\|=1$, corresponding to the two outcomes $By$ and $Bn$, respectively (see Fig.~ \ref{Bloch}).

We assume that the two breakable elastics are parameterized in such a way that the coordinate $x=1$ (resp. $x=-1$) corresponds to the outcome ``yes'' (resp. ``no''), with $x=0$ describing their centers, coinciding also with the center of the Bloch sphere. Each elastic represents a possible measurement, and is described not only by its orientation within the sphere, but also by `the way' it can break. More precisely, considering for instance the elastic associated with measurement $A$, we can describe its breakability by means of a \emph{probability distribution} $\rho_A(x|\psi)$, so that $\int_{x_1}^{x_2}\rho_A(x|\psi)dx$ is the probability that the elastic will break in a point in the interval $[x_1,x_2]$, $-1\leq x_1\leq x_2\leq 1$, in the course of the measurement, when the initial state is $\psi$, and of course we have the unitarity condition $\int_{-1}^{1}\rho_A(x|\psi)dx=1$, expressing the fact that the elastic will break in one of its points, with certainty.

\begin{figure}[!ht]
\centering
\includegraphics[scale =.4]{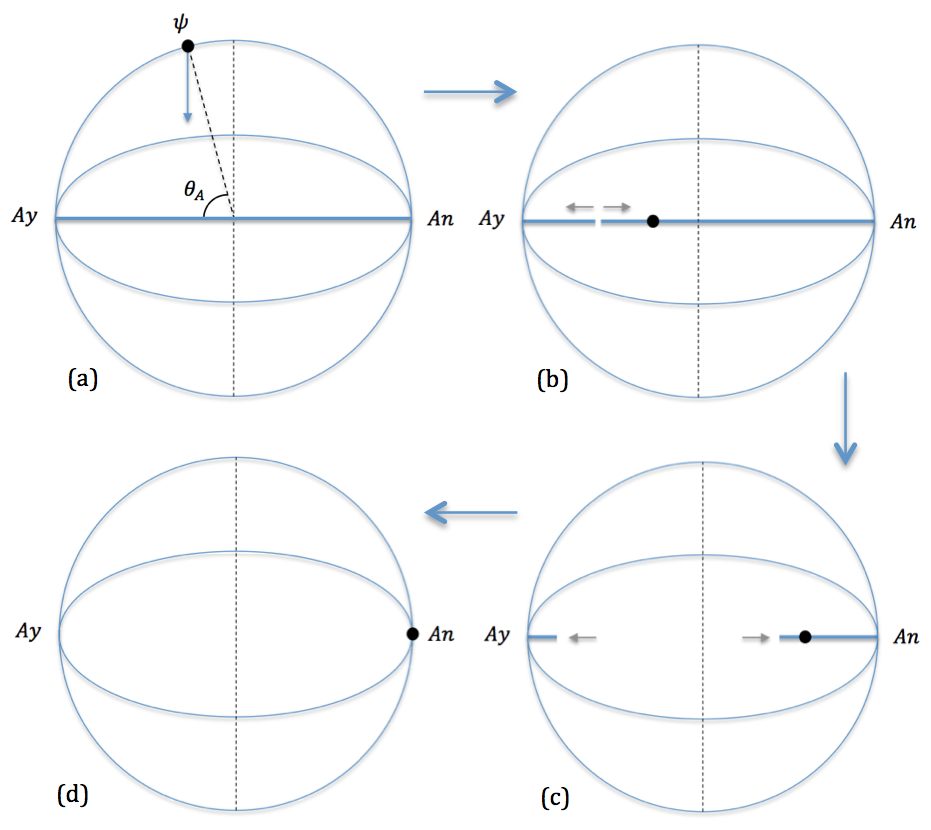}
\caption{The unfolding of a quantum-like $2$-outcome measurement process. Figure (a) shows the abstract point particle, representative of the initial state $\psi$, at the surface of the sphere, in the process of entering into it, by orthogonally ``falling'' onto the elastic band, here describing measurement $A$. In figure (b) the point particle has reached its on-elastic position. The unpredictable breaking point of the elastic is also shown in the figure, which is here assumed to belong to the elastic's fragment going from outcome $Ay$ to the particle position. As a consequence, the contraction of the elastic will draw the point particle to the outcome state $An$, as shown in Figures (c) and (d). 
\label{Measurement}}
\end{figure}

Let us describe more specifically how the quantum-like measurement unfolds (see Fig.~\ref{Measurement}). To the initial (pre-measurement) vector state ${\bf x}_\psi$ we associate an abstract point particle. When the $A$-measurement is executed, a certain probability distribution $\rho_A(x|\psi)$ is
actualized, describing the way the $A$-elastic band will break, in accordance with the fluctuations that are present in that moment in the experimental context. The abstract point particle then plunges into the Bloch sphere, by orthogonally ``falling'' from its surface onto the elastic band, firmly attaching to it (we can consider that the elastic is sticky, or in some way attractive). Then, when the elastic breaks in some point, its two broken fragments will contract toward the corresponding anchor points, bringing with them the point particle. This means that if $x_A$ is the position of the point particle onto the elastic ($x_A= {\bf x}_\psi\cdot {\bf a}_y\equiv\cos\theta_A$), then, if the elastic breaks in a point $\lambda$, with $x_A<\lambda$, the particle will be attached to the elastic fragment collapsing toward ${\bf a}_y$, and will therefore be drawn to that position, which will correspond to the outcome of the measurement. On the other hand, if $x_A>\lambda$, the particle will be attached to the elastic fragment collapsing toward ${\bf a}_n$, and will be drawn to that position (state) at the end of the process. This means that the conditional (transition) probabilities $P(A_y|\psi)=P({\bf x}_\psi\to {\bf a}_y)$ and $P(A_n|\psi)=P({\bf x}_\psi\to {\bf a}_n)$ will be given by the two integrals: 
\begin{equation}
P(A_y|\psi)=\int_{-1}^{\cos\theta_A}\rho_A(x|\psi) dx,\quad P(A_n|\psi)=\int_{\cos\theta_A}^{1}\rho_A(x|\psi) dx.
\label{A-integrals}
\end{equation}

When the breaking point of the elastic exactly coincides with the landing point of the particle, i.e., $\lambda = \cos\theta_A$, we are in a situation of classical unstable equilibrium, and the outcome is not predetermined. However, these exceptional $\lambda$-values are clearly of zero measure, and will not contribute to the determination of the probabilities (\ref{A-integrals}). We also observe that to each breaking point of the elastic (excluding the above mentioned exceptional points) is associated a \emph{deterministic} measurement-interaction between the point particle (representing the measured entity) and the elastic (representing the measuring system, or better, that part of the measuring systems that is relevant for the description of the process in question). In a general measurement context, the deterministic measurement-interactions remain ``hidden,'' in the sense that it is impossible to predict in advance which of them will be actualized, at each run of the measurement. This process of \emph{actualization of a potential, almost deterministic, measurement-interaction}, is what is described in our model by the given of the probability distribution $\rho_A(x|\psi)$, which we assume can generally depend also on the pre-measurement state. 

Before continuing in our analysis of the sequential measurements, it is worth observing that when  $\rho_A(x|\psi)={1\over 2}$, i.e., when the probability distribution is globally uniform, (\ref{A-integrals}) reduces to the well-known Born rule's probability formulae: 
\begin{equation}
P(A_y|\psi)={1\over 2}(1+\cos\theta_A),\quad P(A_n|\psi)={1\over 2}(1-\cos\theta_A).
\label{A-quantum}
\end{equation}
In other terms, the model reduces to standard quantum mechanics when the probability distributions describing the elastic bands become globally uniform. This is a general result, not limited to $2$-outcome measurements,  as can be seen from the more general analysis in 
\citet{AertsSassolideBianchi2014,AertsSassolideBianchi2015a,AertsSassolideBianchi2015b,AertsSassolideBianchi2015c}.

With a similar notation, we can also write the conditional (transition) probabilities $P(B_y|\psi)=P({\bf x}_\psi\to {\bf b}_y)$ and $P(B_n|\psi)=P({\bf x}_\psi\to {\bf b}_n)$, as the integrals:
\begin{equation}
P(B_y|\psi)=\int_{-1}^{\cos\theta_B}\rho_B(x|\psi) dx,\quad P(B_n|\psi)=\int_{\cos\theta_B}^{1}\rho_B(x|\psi) dx,
\label{B-integrals}
\end{equation}
where $x_B= {\bf x}_\psi\cdot {\bf b}_y\equiv\cos\theta_B$ is the landing point of the point particle onto the $B$-elastic band, and $\rho_B(x|\psi)$ is the probability distribution associated with the latter. Proceeding with the same logic, defining $\cos\theta \equiv {\bf a}_y\cdot {\bf b}_y$, we can also write, for the conditional (transition) probabilites $P(By|\psi Ay)=P({\bf a}_y\to {\bf b}_y|{\bf x}_\psi\to {\bf a}_y)$, $P(Bn|\psi Ay)=P({\bf a}_y\to {\bf b}_n|{\bf x}_\psi\to {\bf a}_y)$, $P(By|\psi An)=P({\bf a}_n\to {\bf b}_y|{\bf x}_\psi\to {\bf a}_n)$ and $P(Bn|\psi An)=P({\bf a}_n\to {\bf b}_n|{\bf x}_\psi\to {\bf a}_n)$:
\begin{eqnarray}
&&P(By|\psi Ay)=\int_{-1}^{\cos\theta}\rho_B(x|\psi Ay) dx,\quad P(Bn|\psi Ay)=\int_{\cos\theta}^{1}\rho_B(x|\psi Ay) dx,\nonumber\\
&&P(By|\psi An)=\int_{-1}^{-\cos\theta}\rho_B(x|\psi An) dx,\quad P(Bn|\psi An)=\int_{-\cos\theta}^{1}\rho_B(x| \psi An) dx,
\label{B|A-integrals}
\end{eqnarray}
where $\rho_B(x|\psi Ay)$ (resp., $\rho_B(x|\psi An)$) is the probability distribution actualized during the second $B$-measurement, knowing that the first $A$-measurement produced the transition $\psi \to Ay$ (resp., $\psi \to An$). Inserting (\ref{B|A-integrals}) and (\ref{B-integrals}) into (\ref{conditionalAB}), we thus obtain: 
\begin{eqnarray}
&&P(AyBy|\psi)= \int_{-1}^{\cos\theta}\rho_B(x|\psi Ay) dx \int_{-1}^{\cos\theta_A}\rho_A(x|\psi) dx,\nonumber\\
&&P(AyBn|\psi)= \int_{\cos\theta}^{1}\rho_B(x|\psi Ay) dx \int_{-1}^{\cos\theta_A}\rho_A(x|\psi) dx,\nonumber\\
&&P(AnBy|\psi)= \int_{-1}^{-\cos\theta}\rho_B(x|\psi An) dx \int_{\cos\theta_A}^{1}\rho_A(x|\psi) dx,\nonumber\\
&&P(AnBn|\psi)=\int_{-\cos\theta}^{1}\rho_B(x|\psi An) dx \int_{\cos\theta_A}^{1}\rho_A(x|\psi) dx,
\label{AB-integrals}
\end{eqnarray}
and for the reversed order measurement $G=BA$, we can also write, \emph{mutatis mutandis}: 
\begin{eqnarray}
&&P(ByAy|\psi)= \int_{-1}^{\cos\theta}\rho_A(x|\psi By) dx \int_{-1}^{\cos\theta_B}\rho_B(x|\psi) dx,\nonumber\\
&&P(ByAn|\psi)= \int_{\cos\theta}^{1}\rho_A(x|\psi By) dx \int_{-1}^{\cos\theta_B}\rho_B(x|\psi) dx,\nonumber\\
&&P(BnAy|\psi)= \int_{-1}^{-\cos\theta}\rho_A(x|\psi Bn) dx \int_{\cos\theta_B}^{1}\rho_B(x|\psi) dx,\nonumber\\
&&P(BnAn|\psi)=\int_{-\cos\theta}^{1}\rho_A(x|\psi Bn) dx \int_{\cos\theta_B}^{1}\rho_B(x|\psi) dx.
\label{BA-integrals}
\end{eqnarray}

\section{Three simplifying hypotheses}
\label{simplifying hypotheses}

The expressions (\ref{AB-integrals})-(\ref{BA-integrals}) that we have just derived remain of course very general and can be used to model whatever experimental data coming from two 4-outcome measurements $F$ and $G$, be them the result of two sequential 2-outcome measurements or not. So, what we need now to do is to add, in a gradual way, more structure to the model, by enouncing a certain number of simplifying hypothesis that, when verified,  impose certain constraints on the experimental probabilities. Of course, we  need to add sufficient constraints to make the model solvable, but at the same time we must take care not to add too many constraints, for the model to remain sufficiently general and be able to describe also beyond-quantum structures. In other terms, we have to use in combination Chatton's anti-razor (no less than necessary) and Occam's razor (no more than necessary). We denote our first hypothesis \emph{weak compatibility}. 
\\

\noindent {\bf Weak compatibility}. Two 2-outcome measurements $A$ and $B$ are said to be \emph{weakly compatible}, relative to each other and to the initial state $\psi$, if $\rho_A(x|\psi)=\rho_A(x|\psi By)=\rho_A(x|\psi Bn)\equiv\rho_A(x)$, and $\rho_B(x|\psi)=\rho_B(x|\psi Ay)=\rho_B(x|\psi An)\equiv\rho_B(x)$.
\\

\noindent In other terms, weak compatibility tells us that the fluctuations responsible for the actualization of a measurement-interaction during the execution of the $A$-measurements (resp., the $B$-measurement) are the same if the measurement is performed as a first measurement, on the initial state $\psi$, or following the execution of measurement $B$ (resp., $A$), the latter being always executed on the initial state $\psi$.

Our second hypothesis is meant to reduce the class of functions that are allowed to describe the probability distributions $\rho_A(x)$ and $\rho_B(x)$, to
enable us to explicitly perform the integrals in (\ref{AB-integrals})-(\ref{BA-integrals}). We call this hypothesis \emph{local uniformity}.\footnote{To draw a parallel with a similar situation in physics, think about when a potential barrier of general shape is approximated by a square barrier, in order to explicitly solve the stationary Schr{\oe}dinger equation.} 
\\

\noindent {\bf Local uniformity}. A 2-outcome measurement $A$ (resp. $B$) is said to be \emph{locally uniform} if the way its representative elastic band breaks is described by a probability distribution  $\rho_A(x)$ (resp. $\rho_B(x)$) of the form: 
\begin{equation}
\rho_A(x) = {1\over 2\epsilon_A}
\begin{cases}
0, & x\in [-1, d_A- \epsilon_A) \\
1, & x \in [d_A- \epsilon_A,d_A+ \epsilon_A]\\
0, & x\in (d_A+ \epsilon_A,1],
\end{cases}\quad \rho_B(x) = {1\over 2\epsilon_B}
\begin{cases}
0, & x\in [-1, d_B- \epsilon_B) \\
1, & x \in [d_B- \epsilon_B,d_B+ \epsilon_B]\\
0, & x\in (d_B+ \epsilon_B,1],
\end{cases}
\label{locallyuniform-rho}
\end{equation}
where $\epsilon_A, \epsilon_B\in [0,1]$, $d_A\in [-1+\epsilon_A, 1-\epsilon_A]$, and   $d_B\in [-1+\epsilon_B, 1-\epsilon_B]$.
\\

\noindent In other terms, a locally uniform measurement is characterized by an elastic band that can \emph{uniformly} break only inside a connected internal region, not necessarily centered with respect to the origin of the Bloch sphere. Eq.~(\ref{A-integrals}) can then be explicitly solved, to give: 
\begin{equation}
P(A_y|\psi)=\begin{cases}
0, & \cos\theta_A\in [-1, d_A- \epsilon_A) \\
{1\over 2}(1+{\cos\theta_A -d_A\over \epsilon_A}), & \cos\theta_A \in [d_A- \epsilon_A,d_A+ \epsilon_A]\\
1, & \cos\theta_A\in (d_A+ \epsilon_A,1],
\end{cases}
\label{Ay-integral-local-uniform}
\end{equation}
\begin{equation}
P(A_n|\psi)=\begin{cases}
1, & \cos\theta_A\in [-1, d_A- \epsilon_A) \\
{1\over 2}(1-{\cos\theta_A -d_A\over \epsilon_A}), & \cos\theta_A \in [d_A- \epsilon_A,d_A+ \epsilon_A]\\
0, & \cos\theta_A\in (d_A+ \epsilon_A,1],
\end{cases}
\label{An-integral-local-uniform}
\end{equation}
and similarly for (\ref{B-integrals}). And of course, in the (non-singular) limit $(\epsilon_A,d_A)\to (1,0)$, one finds back the pure quantum formulae (\ref{A-quantum}).

The above two hypothesis are certainly sufficient to write down an exact solution for (\ref{AB-integrals})-(\ref{BA-integrals}). However, it would still be unnecessarily complicated for the purposes of our discussion, which is the reason why we further limit our considerations to situations where a variation of the pre-measurement state, however small, is always able to produce a non-zero variation of the values of the outcome probabilities. We call this property \emph{sensitivity to pre-measurement state}. 
\\

\noindent {\bf Sensitivity to  pre-measurement state}. A 2-outcome measurement $A$ (resp. $B$) is said to be \emph{sensitive to the pre-measurement state} $\psi$, if the outcome probability $P(Ay|\varphi)$ (resp., $P(By|\varphi)$) is not a constant function of the initial state $\varphi$, when the latter is varied in a voisinage of $\psi$.
\\

\noindent Clearly, a locally uniform measurement $A$ (resp., $B$) will be sensitive to a pre-measurement state $\psi$ only if the representative unit vector ${\bf x}_\psi$ is such that: ${\bf x}_\psi\cdot {\bf a}_y\equiv\cos\theta_A\in [d_A- \epsilon_A,d_A+ \epsilon_A]$ (resp., ${\bf x}_\psi\cdot {\bf b}_y\equiv\cos\theta_B\in [d_B- \epsilon_B,d_B+ \epsilon_B]$), i.e., if the point particle representative of the pre-measurement state orthogonally ``falls'' onto the internal, uniformly breakable part of the elastic, and not on the external unbreakable one (quantum measurements, characterized by globally uniformly elastics, are of course  sensitive to all pre-measurement states).

\section{Explicitly solving the model}
\label{solving the model}

In the following, we assume that the 2-outcome measurements $A$ and $B$, forming the two 4-outcome sequential measurements $F=AB$ and $G=BA$, are weakly compatible, locally uniform and sensitive to all the pre-measurement states intervening during the execution of the two sequences of measurements. This will be the case if the following $4$ constraints are verified: 
\begin{equation}
\cos\theta, \cos\theta_A \in [d_A- \epsilon_A,d_A+ \epsilon_A], \quad \cos\theta, \cos\theta_B \in [d_B- \epsilon_B,d_B+ \epsilon_B].
\label{generalconstraints}
\end{equation}
Indeed, if this is the case, the point particle will always orthogonally project onto the internal breakable segment of both the $A$-elastic and $B$-elastic, either in the first measurement or in the second one. Then, for the $AB$ sequential measurement, we have: 
\begin{eqnarray}
&&P(AyBy|\psi)={1\over 4}(1+{\cos\theta -d_B\over \epsilon_B})(1+{\cos\theta_A -d_A\over \epsilon_A}), \nonumber\\
&&P(AyBn|\psi)={1\over 4} (1-{\cos\theta -d_B\over \epsilon_B})(1+{\cos\theta_A -d_A\over \epsilon_A}),\nonumber\\
&&P(AnBn|\psi)={1\over 4} (1+{\cos\theta +d_B\over \epsilon_B})(1-{\cos\theta_A -d_A\over \epsilon_A}),\nonumber\\
&&P(AnBy|\psi)={1\over 4} (1-{\cos\theta +d_B\over \epsilon_B})(1-{\cos\theta_A -d_A\over \epsilon_A}),
\label{explicitsolutionAB}
\end{eqnarray}
and similarly, the $BA$ sequential measurement gives:
\begin{eqnarray}
&&P(ByAy|\psi)={1\over 4}(1+{\cos\theta -d_A\over \epsilon_A})(1+{\cos\theta_B -d_B\over \epsilon_B}), \nonumber\\
&&P(ByAn|\psi)={1\over 4} (1-{\cos\theta -d_A\over \epsilon_A})(1+{\cos\theta_B -d_B\over \epsilon_B}),\nonumber\\
&&P(BnAn|\psi)={1\over 4} (1+{\cos\theta +d_A\over \epsilon_A})(1-{\cos\theta_B -d_B\over \epsilon_B}),\nonumber\\
&&P(BnAy|\psi)={1\over 4} (1-{\cos\theta +d_A\over \epsilon_A})(1-{\cos\theta_B -d_B\over \epsilon_B}).
\label{explicitsolutionBA}
\end{eqnarray}
From the first two equations in (\ref{explicitsolutionAB}), we obtain: 
\begin{equation}
{P(AyBy|\psi)\over P(AyBn|\psi)}={ 1+{\cos\theta -d_B\over \epsilon_B}\over 1-{\cos\theta -d_B\over \epsilon_B}},
\end{equation}
from which we deduce that:
\begin{equation}
{d_B\over \epsilon_B} = {\cos\theta\over \epsilon_B} - {P(AyBy|\psi)-P(AyBn|\psi)\over P(AyBy|\psi)+P(AyBn|\psi)}.
\label{ybfirst}
\end{equation}
Doing the same with the last two equations in (\ref{explicitsolutionAB}), we also find: 
\begin{equation}
{d_B\over \epsilon_B} = -{\cos\theta\over \epsilon_B} + {P(AnBn|\psi)-P(AnBy|\psi)\over P(AnBn|\psi)+P(AnBy|\psi)}.
\label{ybsecond}
\end{equation}
Combining (\ref{ybfirst}) and (\ref{ybsecond}), we obtain: 
\begin{eqnarray}
{\cos\theta\over \epsilon_B} &=& {1\over 2} \left[{P(AnBn|\psi)-P(AnBy|\psi)\over P(AnBn|\psi)+P(AnBy|\psi)} +{P(AyBy|\psi)-P(AyBn|\psi)\over P(AyBy|\psi)+P(AyBn|\psi)}\right],\nonumber\\
{d_B\over \epsilon_B} &=& {1\over 2} \left[{P(AnBn|\psi)-P(AnBy|\psi)\over P(AnBn|\psi)+P(AnBy|\psi)} -{P(AyBy|\psi)-P(AyBn|\psi)\over P(AyBy|\psi)+P(AyBn|\psi)}\right].
\label{cos-d-B}
\end{eqnarray}
Considering also that $P(Ay|\psi) = {1\over 2}(1+{\cos\theta_A -d_A\over \epsilon_A})$, and $P(Ay|\psi)=P(AyBy|\psi) + P(AyBn|\psi)$, we have: 
\begin{equation}
{\cos\theta_A \over \epsilon_A}-{d_A\over \epsilon_A}=2[P(AyBy|\psi) + P(AyBn|\psi)]-1.
\label{difference-A}
\end{equation}
The same calculation for the $BA$ measurement yields: 
\begin{eqnarray}
{\cos\theta\over \epsilon_A} &=& {1\over 2} \left[{P(BnAn|\psi)-P(BnAy|\psi)\over P(BnAn|\psi)+P(BnAy|\psi)} +{P(ByAy|\psi)-P(ByAn|\psi)\over P(ByAy|\psi)+P(ByAn|\psi)}\right],\nonumber \\
{d_A\over \epsilon_A} &=& {1\over 2} \left[{P(BnAn|\psi)-P(BnAy|\psi)\over P(BnAn|\psi)+P(BnAy|\psi)} -{P(ByAy|\psi)-P(ByAn|\psi)\over P(ByAy|\psi)+P(ByAn|\psi)}\right],
\label{cos-d-A}
\end{eqnarray}
as well as: 
\begin{equation}
{\cos\theta_B \over \epsilon_B}-{d_B\over \epsilon_B}=2[P(ByAy|\psi) + P(ByAn|\psi)]-1,
\label{difference-B}
\end{equation}
and inserting the second equation in (\ref{cos-d-A}) into (\ref{difference-A}), and the second equation in (\ref{cos-d-B}) into (\ref{difference-B}), explicit expressions for ${\cos\theta_A \over \epsilon_A}$ and ${\cos\theta_B \over \epsilon_B}$ can also be obtained.

It is important to observe that the model's system of equations is underdetermined, in the sense that the $8$ outcome probabilities can determine all the parameters but one. This means that we are free to choose one of the parameters, for instance $\epsilon_A$, and by doing so all the others will be automatically determined. When choosing a value for $\epsilon_A$, we have however to remember that, in accordance with (\ref{locallyuniform-rho}), it needs to obey the constraint: $\epsilon_A + d_A\leq 1$, as is clear that the uniformly breakable region of the elastic cannot extend outside of the Bloch sphere. This means that: $\epsilon_A(1+{d_A\over \epsilon_A})\leq 1$, giving: 
\begin{equation}
\epsilon_A\leq {1\over 1+{d_A\over \epsilon_A}}.
\label{constraint-on-epsilon}
\end{equation}

It immediately follows from (\ref{constraint-on-epsilon}) that when the experimental probabilities are such that the right hand side of the second equality in (\ref{cos-d-A}) is different from zero, then the data cannot be modeled by quantum mechanics, but require a more general class of measurements. The same is true of course when the right hand side of the second equality in (\ref{cos-d-B}) is also different from zero, as is clear that a pure quantum modeling requires both elastic bands to be globally (and not just locally) uniformly breakable.

\section{Modeling Moore's data}
\label{modeling the data}

In this section we use the above explicit solution, obtained under the three simplifying hypothesis of weak compatibility, local uniformity and sensitivity to pre-measurement state, to model some of the data obtained in a Gallup poll conducted in 1997, as presented in a review of question order effects by \citet{Moore2002}. More precisely, we will use the probabilities given by \citet{WangBusemeyer2013} (see also \citep{WangEtal2014}), which are different from those in \citep{Moore2002}, as the authors have rightly excluded from the statistics those respondents who did not provided a ``yes'' or ``no'' answer (in other terms, they did not include in the total count the ``don't know'' responses).

\subsection{Clinton/Gore}

In one of the experiments, a thousand respondents were subjected to two interrogative contexts, consisting of a pair of questions asked in a sequence. The first question, let us denote it $A$, is: ``Do you generally think Bill Clinton is honest and trustworthy?'' The second question, let us denote it $B$, is: ``Do you generally think Al Gore is honest and trustworthy?'' Half of the subjects were subjected to the two questions in the order $AB$ (first ``Clinton'' then ``Gore'') and the other half in the reversed order $BA$ (first ``Gore'' then ``Clinton''). The experimental outcome probabilities are:\footnote{Because of a rounding error, the probabilities given in \cite{WangBusemeyer2013} do not exactly sum to $1$, but to $0.9999$, in the $AB$ measurement, and to $1.0001$, in the $BA$ measurement. Since our solution requires the probabilities to exactly sum to $1$, we have corrected the value of $P_{\rm C/G}(AnBn)$, from $0.2886$ to $0.2887$, and the value of $P_{\rm C/G}(BnAn)$, from $0.2130$ to $0.2129$, restoring in this way the exact unitarity of the probability measure.}
\begin{eqnarray}
&P_{\rm C/G}(AyBy)=0.4899,\, P_{\rm C/G}(AyBn)= 0.0447,\, P_{\rm C/G}(AnBy)=0.1767,\, P_{\rm C/G}(AnBn)= 0.2887,\nonumber\\
&P_{\rm C/G}(ByAy)=0.5625,\, P_{\rm C/G}(ByAn) = 0.1991,\, P_{\rm C/G}(BnAy) = 0.0255,\, P_{\rm C/G}(BnAn) = 0.2129.
\label{Prob-Clinton-Gore}
\end{eqnarray}
The above data clearly show question order effects, considering that the probabilities in each of the $4$ columns in (\ref{Prob-Clinton-Gore}) are sensibly different. To investigate the mathematical structure of these effects, we now insert the above values in (\ref{cos-d-B}). This gives: 
\begin{eqnarray}
{\cos\theta\over \epsilon_B} &=& {1\over 2} \left[{0.2887-0.1767\over 0.2887+0.1767} +{0.4899-0.0447\over 0.4899+0.0447}\right]= {1112797\over 2073357},\label{cos-d-B-first-eq-data}\nonumber\\
{d_B\over \epsilon_B} &=& {1\over 2} \left[{0.2887-0.1767\over 0.2887+0.1767} -{0.4899-0.0447\over 0.4899+0.0447}\right]=-{613837\over 2073357},
\label{cos-d-B-data}
\end{eqnarray}
whereas (\ref{difference-A}) gives: 
\begin{equation}
{\cos\theta_A \over \epsilon_A}-{d_A\over \epsilon_A}=2(0.4899+0.0447)-1=0.0692={173\over 2500}.
\label{difference-A-data}
\end{equation}
Doing the same with (\ref{cos-d-A}), we obtain: 
\begin{eqnarray}
{\cos\theta\over \epsilon_A} &=& {1\over 2} \left[{0.2129-0.0255\over 0.2129+0.0255} +{0.5625-0.1991\over 0.5625+0.1991}\right]= {716745\over 1134784},\nonumber\\
{d_A\over \epsilon_A} &=& {1\over 2} \left[{0.2129-0.0255\over 0.2129+0.0255} -{0.5625-0.1991\over 0.5625+0.1991}\right]={175279\over 1134784},
\label{cos-d-A-data}
\end{eqnarray}
and from (\ref{difference-B}), we have: 
\begin{equation}
{\cos\theta_B \over \epsilon_B}-{d_B\over \epsilon_B}=2[0.5625 + 0.1991]-1=0.5232={327\over 625}.
\label{difference-B-data}
\end{equation}
From (\ref{difference-A-data}) and (\ref{cos-d-A-data}), we also find: 
\begin{equation}
{\cos\theta_A \over \epsilon_A}={173\over 2500}+{175279\over 1134784}={158628783\over 709240000},
\label{costhetaAoverepsilonA}
\end{equation}
and (\ref{difference-B-data}) and (\ref{cos-d-B-data}) give:
\begin{equation}
{\cos\theta_B \over \epsilon_B}={327\over 625}-{613837\over 2073357}={294339614\over 1295848125}.
\label{costhetaBoverepsilonB}
\end{equation}
In summary, writing the obtained exact values in approximate form, to facilitate their reading, we have:
\begin{eqnarray}
&&{d_A\over \epsilon_A}\approx 0.15\quad {\cos\theta_A\over \epsilon_A}\approx 0.22,\quad {\cos\theta \over \epsilon_A}\approx 0.63,\nonumber\\
&&{d_B\over \epsilon_B}\approx -0.29,\quad {\cos\theta_B\over \epsilon_B}\approx 0.23,\quad {\cos\theta \over \epsilon_B}\approx 0.54.
\end{eqnarray}

At this point, we need to choose a specific value for $\epsilon_A$, thus allowing us to determine the values of all the other parameters and obtain an exact explicit determination of the experimental probabilities. Before doing that, it is worth observing that two identical elastics ($\epsilon_A=\epsilon_B$, $d_A=d_B$) would clearly be unable to model the data, and that none of the two elastics can be symmetric ($d_A=0$ and/or $d_B=0$). Also, their breakable region cannot be of the same size ($\epsilon_A=\epsilon_B=\epsilon$). In other terms, the two measurements' elastic bands have to be very different and non-symmetric, which means that the structure of the probabilistic data is irreducibly non-Hilbertian (we recall that the Hilbertian structure is recovered when the elastics are all globally uniform). 
 
In view of (\ref{constraint-on-epsilon}), we have: 
\begin{equation}
\epsilon_A\leq {1\over 1+{175279\over 1134784}}={1134784\over 1310063}\approx 0.87.
\label{constraint-on-epsilonA-data}
\end{equation}
Considering that ${d_B\over \epsilon_B}$ is negative, for $\epsilon_B$ we have to consider the constraint: $ d_B -\epsilon_B \geq -1$. This means that: $\epsilon_B(1-{d_B\over \epsilon_B})\leq 1$, i.e., $\epsilon_B\leq {1\over 1-{d_B\over \epsilon_B}}$, giving:
\begin{equation}
\epsilon_B\leq {1\over 1+{613837\over 2073357}}={2073357\over 2687194}\approx 0.77.
\label{constraint-on-epsilonB-data}
\end{equation}
These maximum values are clearly very different form the values $\epsilon_A=\epsilon_B=1$, characterizing quantum mechanics. For simplicity, we choose $\epsilon_A={1\over 2}$. We then obtain the exact values:
\begin{eqnarray}
&&\epsilon_A={1\over 2},\quad \epsilon_B= {1486068262965\over 2525568461696}\nonumber\\
&&d_A = {175279\over 2269568} ,\quad d_B=-{62852085795\over 360795494528} \nonumber\\
&&\cos\theta = {716745\over 2269568} ,\quad \cos\theta_A= {158628783\over 1418480000} ,\quad \cos\theta_B={21096644663643\over 157848028856000},
\label{parameters-values-Clinton}
\end{eqnarray}
or in approximate form: 
\begin{eqnarray}
&&\epsilon_A=0.5,\quad \epsilon_B\approx 0.59\nonumber\\
&&d_A \approx 0.08,\quad d_B\approx-0.17\nonumber\\
&&\cos\theta \approx 0.32,\quad \cos\theta_A\approx 0.11,\quad \cos\theta_B\approx 0.13.
\label{approx-parameters-Clinton}
\end{eqnarray}

We can check that (\ref{parameters-values-Clinton})-(\ref{approx-parameters-Clinton}) obey (\ref{generalconstraints}). This is clearly the case considering that (using the approximate values, for clarity): $[d_A- \epsilon_A,d_A+ \epsilon_A] \approx [-0.42, 0.58]$, which contains the values $0.32$ and $0.04$ of $\cos\theta$ and $\cos\theta_A$, respectively. Also, $[d_B- \epsilon_B,d_B+ \epsilon_B]\approx [-0.76, 0.42]$, which also contains the values $0.32$ and $0.13$ of $\cos\theta$ and $\cos\theta_B$, respectively. A last point we need to check, to be sure that our solution is consistent, is that it is always possible to find a 3-dimensional unit vectors ${\bf x}_\psi$, representing the initial state in the Bloch sphere, which can make the angles $\cos\theta_A$ and $\cos\theta_B$, when projected onto the two unit vectors ${\bf a}_y$ and ${\bf b}_y$, corresponding to the yes-outcomes of the $A$ and $B$ measurements, respectively, with ${\bf a}_y\cdot{\bf b}_y =\cos\theta$. This is indeed the case, as we can always choose:
\begin{equation}
{\bf a}_{y}=(1,00),\quad {\bf b}_{y}=(\cos\theta,\sin\theta,0),\quad {\bf x}_\psi=(\cos\theta_A,{\cos\theta_B-\cos\theta\cos\theta_A\over \sin\theta},x_3),
\end{equation}
where $x_3$ is such that: $\|{\bf x}\|=1$.

\begin{figure}[!ht]
\centering
\includegraphics[scale =.2]{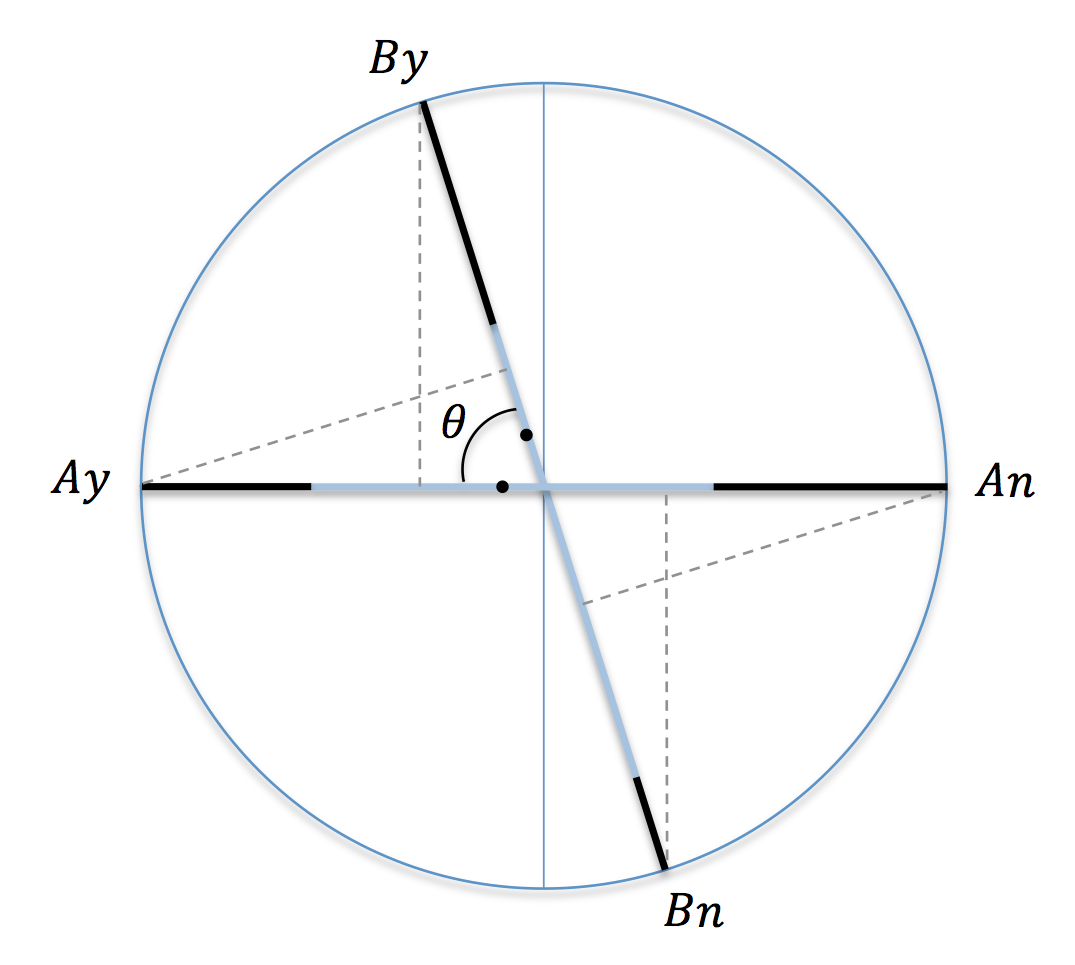}
\caption{Two non-uniform and non-symmetric elastic bands describing the data of the Clinton/Gore experiment. Their relative angle is $\theta \approx 71^\circ$. The initial state vector ${\bf x}_\psi$ is not represented in the drawing, being not located on the same plane of the two elastic bands (see Fig.\ref{Bloch}). Only its projection points onto the two elastic bands are represented (the two black dots). We can observe that the end points of the two elastics orthogonally project onto the internal breakable region of the other elastic, in accordance with the `sensitivity to pre-measurement state' hypothesis. 
\label{ABelastics-Clinton-Gore}}
\end{figure}

Figure~\ref{ABelastics-Clinton-Gore} illustrates our modeling of the Clinton/Gore data, for $\epsilon_A={1\over 2}$. Of course, by considering different choices for the free-parameter $\epsilon_A$, different equivalent representations will be obtained. However, structurally speaking, they will all be very similar.
The two black dots in Fig.~\ref{ABelastics-Clinton-Gore} indicate the two positions of the particle representative of the initial state $\psi$, when it lands 
onto the two elastic bands. We see that the on-elastic position for the $A$-elastic is almost in the middle of its breakable region, whereas the on-elastic position for the $B$-elastic is much closer to the border of the breakable region, towards the $By$ outcome. This is in accordance with the fact that the probability of answering ``yes'' to the Clinton question (the $A$-measurement), $P_{\rm C/G}(Ay)= P_{\rm C/G}(AyBy) + P_{\rm C/G}(AyBn)\approx 0.53$, irrespective of the answer given to the subsequent Gore question (what is called the \emph{non-comparative} context by \citet{Moore2002}), is close to one-half, whereas the probability of answering ``yes'' to the Gore question (the $B$-measurement), $P_{\rm C/G}(By)= P_{\rm C/G}(ByAy)+ P_{\rm C/G}(ByAn)\approx 0.76$, irrespective of the answer to the subsequent Clinton's question, is much higher (with a significant difference of $0.23$). In other terms, the position difference of the point particle on the elastics' breakable regions is what takes into account this difference between $P_{\rm C/G}(Ay)$ and $P_{\rm C/G}(By)$. On the other hand, the relative orientation of the two elastics (the angle $\theta$), the extension and displacement of their locally uniform breakable regions (defined by their ``$\epsilon$-$d$'' parameters), is what accounts for the subtle order effects that are observed, which are too complex to be modeled by only using two globally uniform elastic bands, i.e., two pure quantum measurements.

\subsection{Rose/Jackson}

In another experiment reported by Moore, always performed on a thousand respondents, the interrogative contexts consisted in a pair of questions about the baseball players Pete Rose and Shoeless Joe Jackson. More precisely, the $A$ question was: ``Do you think Rose should or should not be eligible for admission to the Hall of Fame?'' Similarly, the $B$ question was: ``Do you think Jackson should or should not be eligible for admission to the Hall of Fame?'' The obtained experimental probabilities are (also in this case, we use the revisited probability data presented in \citep{WangBusemeyer2013,WangEtal2014}):
\begin{eqnarray}
&&P_{\rm R/J}(AyBy)=0.3379,\, P_{\rm R/J}(AyBn)= 0.3241,\, P_{\rm R/J}(AnBy) =0.0178,\, P_{\rm R/J}(AnBn) = 0.3202,\nonumber\\
&&P_{\rm R/J}(ByAy) =0.4156,\, P_{\rm R/J}(ByAn) = 0.0671,\, P_{\rm R/J}(BnAy) = 0.1234,\, P_{\rm R/J}(BnAn) = 0.3939.
\label{Prob-Rose-Jackson}
\end{eqnarray}
This time we have: 
\begin{eqnarray}
{\cos\theta\over \epsilon_B} &=& {1\over 2} \left[{0.3202-0.0178\over 0.3202+0.0178} +{0.3379-0.3241\over 0.3379+0.3241}\right]={512133\over 1118780},\nonumber\\
{d_B\over \epsilon_B} &=& {1\over 2} \left[{0.3202-0.0178\over 0.3202+0.0178} -{0.3379-0.3241\over 0.3379+0.3241}\right]= {488811\over 1118780},\nonumber\\
{\cos\theta\over \epsilon_A} &=& {1\over 2} \left[{0.3939-0.1234\over 0.3939+0.1234} +{0.4156-0.0671\over 0.4156+0.0671}\right]= {15542470\over 24970071},\nonumber\\
{d_A\over \epsilon_A} &=& {1\over 2} \left[{0.3939-0.1234\over 0.3939+0.1234} -{0.4156-0.0671\over 0.4156+0.0671}\right]= -{2485435\over 24970071},
\end{eqnarray}
and from the two equalities:
\begin{eqnarray}
{\cos\theta_A \over \epsilon_A}-{d_A\over \epsilon_A}&=&2(0.3379+0.3241)-1=0.324={81\over 250},\nonumber\\
{\cos\theta_B \over \epsilon_B}-{d_B\over \epsilon_B}&=&2(0.4156+0.0671)-1=0346={173\over 5000},
\end{eqnarray}
we obtain: 
\begin{equation}
{\cos\theta_A\over \epsilon_A}= {1401217001\over 6242517750},\quad {\cos\theta_B\over \epsilon_B}= {112525303\over 279695000}.
\end{equation}
Writing the obtained values in approximate form, we thus have:
\begin{eqnarray}
&&{d_A\over \epsilon_A}\approx -0.10 , \quad {\cos\theta_A\over \epsilon_A}\approx 0.22,\quad {\cos\theta \over \epsilon_A}\approx 0.62, \nonumber\\
&&{d_B\over \epsilon_B}\approx 0.44,\quad {\cos\theta_B\over \epsilon_B}\approx 0.40,\quad {\cos\theta \over \epsilon_B}\approx 0.46.
\end{eqnarray}
As we did for the previous data, we choose $\epsilon_A={1\over 2}$. It is then easy to check that all the constraints are obeyed, and one finds: 
\begin{eqnarray}
&&\epsilon_A={1\over 2},\quad \epsilon_B= {8694302293300\over 12787997371443},\nonumber\\
&&d_A = -{2485435\over 49940142},\quad d_B= {1266221717195\over 4262665790481},\nonumber\\
&&\cos\theta = {7771235\over 24970071} ,\quad \cos\theta_A= {1401217001\over 12485035500},\quad \cos\theta_B= {174892114611841\over 639399868572150},
\end{eqnarray}
or in approximate form: 
\begin{eqnarray}
&&\epsilon_A=0.5,\quad \epsilon_B\approx 0.68\nonumber\\
&&d_A \approx -0.05,\quad d_B\approx 0.30\nonumber\\
&&\cos\theta \approx 0.31 ,\quad \cos\theta_A\approx 0.11,\quad \cos\theta_B \approx 0.27.
\end{eqnarray}

\begin{figure}[!ht]
\centering
\includegraphics[scale =.2]{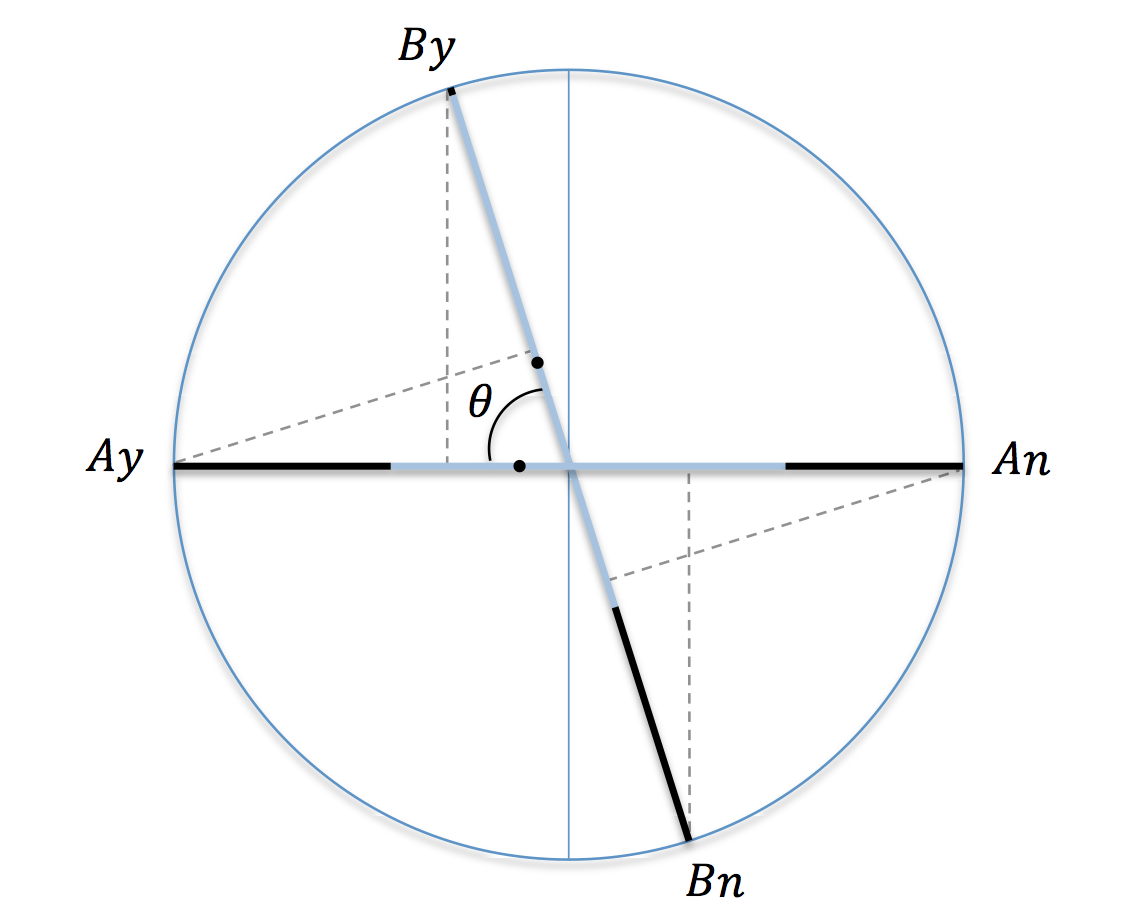}
\caption{Two non-uniform and non-symmetric elastic bands describing the data of the Rose/Jackson experiment. Their relative angle is $\theta \approx 72^\circ$. The initial state vector ${\bf x}_\psi$ is not represented in the drawing, being not located on the same plane of the two elastic bands (see Fig.\ref{Bloch}). Only its projection points onto the two elastic bands are represented (the two black dots). We can observe that the end points of the two elastics orthogonally project onto the internal breakable region of the other elastic, 
in accordance with the `sensitivity to pre-measurement state' hypothesis.
\label{ABelastics-Rose-Jackson}}
\end{figure}

\section{Quantum tests}
\label{quantum tests}

In the previous sections we have provided an exact modeling of some paradigmatic data exhibiting question order effects, using a mathematical formalism that is more general than the standard formalism of quantum mechanics. If on one hand it was to be expected that the data could not be exactly modeled by the Born rule, on the other hand it may come as a little surprise that the structure of the Clinton/Gore probabilities are not only irreducibly non-Hilbertian, but also not at all close to a Hilbertian model. Indeed, the Clinton/Gore data are structurally similar to the Rose/Jackson data, despite of the fact that they are generally considered to be very different, considering that there is a test, expressed by a so-called \emph{QQ-equality}, which is passed (although not in an exact way) by the Clinton/Gore data, but not by the Rose/Jackson data \citep{WangBusemeyer2013, WangEtal2014}.

The explanation that was given for this difference, is that the validity of the QQ-equality in psychological order effect measurements would depend on the critical assumption that, quoting from \citet{WangBusemeyer2013}: ``the only factor that changes the context or the state for answering a question is answering its preceding question.'' In the Rose/Jackson data this assumption is believed to be violated, as during the survey the subjects were provided with some background information. For instance, before the Pete Rose question, the following information was typically given: ``As you may know, former Major League player Pete Rose is ineligible for baseball’s Hall of Fame due to charges that he had gambled on baseball games.'' Information was also provided before addressing the Joe Jackson question, typically: ``As you may know, Shoeless Joe Jackson is ineligible for baseball’s Hall of Fame due to charges that he took money from gamblers in exchange for fixing the 1919 World Series.'' 

This means that, quoting again from \citet{WangBusemeyer2013}: ``the initial belief state when Rose/Jackson was asked first was affected by the information provided about the particular player. Furthermore, the context for the second question was changed not only by answering the first question but also sequentially by the additional background information on the player in the second question. [...] From the perspective of the QQ model, actually the two orders in the data set were Rose background-Rose question-Jackson background-Jackson question versus Jackson background-Jackson question-Rose background-Rose question.''

According to \citet{WangBusemeyer2013}, this would be sufficient to explain why the QQ-equality is (almost) obeyed by the Clinton/Gore data but manifestly disobeyed by the Rose/Jackson data. But this is quite surprising, considering that it is always possible to incorporate the background information into the questions themselves, for instance defining the following modified interrogative contexts: ``As you may know, former Major League player Pete Rose is ineligible for baseball’s Hall of Fame due to charges that he had gambled on baseball games; do you think Rose should or should not be eligible for admission to the Hall of Fame?'' and: ``As you may know, Shoeless Joe Jackson is ineligible for baseball’s Hall of Fame due to charges that he took money from gamblers in exchange for fixing the 1919 World Series; do you think Jackson should or should not be eligible for admission to the Hall of Fame?''

So, we now have two rephrased questions, with no background information provided anymore (as it became part of the question), and one would expect the QQ-equality to also be (almost) obeyed in this case. There is undoubtedly something that needs to be clarified here, not only as regards the true content of the QQ-equality (is it really a quantum test?), but also the proper way to interpret the quantum formalism, and its GTR-model generalization, when used to model cognitive situations. 

Let us have a closer look to the QQ-equality, derived by \citet{WangBusemeyer2013} as part of their QQ-model, where the subjects' beliefs are represented by state vectors in a Hilbert space and the interrogative contexts by generally non commuting orthogonal projection operators. Using the property of the scalar product (or of the trace, in the more general setting where density matrices are also allowed to represent states), it is possible to demonstrate the following quantum mechanical equality \citep{BusemeyerBruza2012,WangBusemeyer2013}: 
\begin{equation}
q\equiv [P(AyBn|\psi)+P(AnBy|\psi)]-[P(ByAn|\psi)+ P(BnAy|\psi)]=0,
\label{QQ-equality}
\end{equation}
which by using (\ref{unitarity2}), can also be written in the form: 
\begin{equation}
q\equiv [P(ByAy|\psi) - P(AyBy|\psi)]+[P(BnAn|\psi) -P(AnBn|\psi)]=0.
\label{QQ-equality2}
\end{equation}
Inserting the explicit solutions (\ref{explicitsolutionAB})-(\ref{explicitsolutionBA}) into (\ref{QQ-equality2}), one finds, after some simple algebra: 
\begin{equation}
q= {\cos\theta \over 2}\left({1\over \epsilon_A}-{1\over \epsilon_B}\right)-{1\over 2}\left({d_A\over\epsilon_A}{\cos\theta_B\over \epsilon_B}-{d_B\over\epsilon_B}{\cos\theta_A\over \epsilon_A}\right).
\label{QQ-equality3}
\end{equation}

As we explained, the pure quantum situation is recovered when the elastic bands become globally uniform, i.e., in the limit $(\epsilon_A,d_A), (\epsilon_B,d_B)\to (1,0)$, for which, evidently, $q\to 0$. This is so because in this limit both terms in the difference (\ref{QQ-equality3}) tend to zero. The first term quantify the difference in terms of the extension of the ``potentiality region'' characterizing the two measurements, i.e., the difference in terms of the number of hidden measurement-interactions that are available to be actualized (characterizing, in a sense, the amount of indeterminism expressed by the interrogative context). The second term, instead, is more articulated, and expresses the relative asymmetry of the two measurements, responsible for part of the observed order effects. For simplicity, let us call the first term the ``relative indeterminism'' and the second term the ``relative asymmetry'' contribution.

Inserting into (\ref{QQ-equality3}) the values (\ref{parameters-values-Clinton}) of the parameters for the Clinton/Gore data, we find:\footnote{Note that in \citep{WangBusemeyer2013} the slightly different value $q=-0.0031$ was given, as the probability data they have used do not exactly sum up to 1; see the previous footnote.} 
\begin{equation}
q= \underbrace{223284032117\over 4705624699776}_{\approx 0.0474} - \underbrace{148963769472677\over 2941015437360000}_{\approx 0.0506}=-{6\over 625}=-0.0032.
\label{QQ-equality-Clinton}
\end{equation}
We immediately see in (\ref{QQ-equality-Clinton}) that the reason why the Clinton/Gore probabilities (almost) obey (\ref{QQ-equality2}) is very different from the reason why the equality is obeyed by quantum probabilities. Here $q$ is almost zero because the ``relative indeterminism'' and the ``relative asymmetry'' contributions almost perfectly cancel each other, and not because they are both individually close to zero, as it should be the case in a close to quantum situation. This almost perfect compensation is certainly the expression of a remarkable symmetry of the probabilistic data, but such symmetry is not characteristic only of Hilbertian probability models. 

Let us also consider the $q$-value for the Rose/Jackson data:
\begin{equation}
q=\underbrace{4600607215157\over 55872032066760}_{\approx 0.0823} - \underbrace{\left[-{60287788121101\over 873000501043125}\right]}_{\approx -0.0691} = {757\over 5000} = 0.1514.
\end{equation}
We see that the ``relative indeterminism'' and the ``relative asymmetry'' contributions are of the same ``size'' than those of the Clinton/Gore data, the difference being here that they have opposite sign, and therefore cannot compensate each other. We also observe that the maximum absolute value $|q^{\rm max}|$ that can be taken by $q$ in a general sequential measurement is $1$, which means that in the Clinton/Gore experiment we have that $|q|$ is $0.32\%$ of $|q^{\rm max}|$, whereas in the Rose/Jackson experiment $|q|$ is $15\%$ of $|q^{\rm max}|$. 

To make more stringent our point that the Clinton/Gore data are decidedly non-Hilbertian, let us use our solution to derive additional relations that a quantum system must obey, but that are flagrantly disobeyed by the Clinton/Gore data. When $\epsilon_A=1$ and $d_A = 0$, the right hand side of the second equation in (\ref{cos-d-A}) has to be zero. This is the case if and only if: 
\begin{eqnarray}
\lefteqn{[P(BnAn|\psi)-P(BnAy|\psi)][P(ByAy|\psi)+P(ByAn|\psi)]}\\
&&=[P(BnAn|\psi)+P(BnAy|\psi)][P(ByAy|\psi)-P(ByAn|\psi)].
\label{constraint-on-probabilities}
\end{eqnarray}
Simplifying the above expression, we thus obtain: 
\begin{equation}
q_1\equiv P(ByAn|\psi)P(BnAn|\psi)-P(BnAy|\psi)P(ByAy|\psi)=0.
\label{q1}
\end{equation}
Reasoning in the same way with the $B$-measurement, (\ref{cos-d-B}) yields the equality:
\begin{equation}
q_2\equiv P(AyBn|\psi)P(AnBn|\psi)-P(AnBy|\psi)P(AyBy|\psi)=0.
\label{q2}
\end{equation}
Also, when $\epsilon_A=\epsilon_B = 1$, the right hand sides of the first equations in (\ref{cos-d-B}) and (\ref{cos-d-A}) have to coincide. After a simple calculation, this provides the additional equality: 
\begin{equation}
q_3\equiv P(AnBy|\psi)P(BnAn|\psi)-P(AnBn|\psi)P(BnAy|\psi) =0,
\label{q3}
\end{equation}
and of course, by combining together the above three equalities, additional relations can be found. 

The maximum absolute values that can be taken by these three quantities are $|q_1^{\rm max}|=|q^{\rm max}_2|=0.25$, and $|q_3^{\rm max}|=1$. If we calculate the values of $q_{1}$, $q_2$ and $q_3$, for the Clinton/Gore data, we find $q_1\approx 0.028$, $q_2\approx -0.073$ and $q_3\approx 0.029$. We thus see that $|q_1|$, $|q_2|$ and $|q_3|$ are now respectively $11.2\%$, $29.2\%$ and $2.9\%$ of the maximum values, which clearly represent strong violations of the quantum predictions. So, we can say that it is almost by a fortuitous circumstance that the Clinton/Gore probabilities (almost) obey (\ref{QQ-equality}), as the same probabilities violate in a very flagrant way other critical quantum (parameter-free) equalities.

A similar conclusion was recently reached by \citet{Boyer-KassemEtal2015}, who by reasoning directly with the notion of conditional probabilities also pointed out the insufficiency of the QQ-equality in characterizing a (non-degenerate) quantum probability model. Indeed, as we have also seen in the previous sections, in the quantum formalism transition probabilities from one state to another state are expressions of conditional statements, and because of the conjugate symmetry of the inner product the probability of a transition will not depend on its order (a property requiring the elastics to be globally uniform). This introduces a symmetry also in the conditional probabilities, that \citet{Boyer-KassemEtal2015} organize in a set of equations that
they call the Grand Reciprocity equations, which need also to be satisfied by all quantum models, as the three equations (\ref{q1})-(\ref{q3}) and the QQ-equation, need to be, but are not satisfied by most of the existing data sets. 

We therefore see that a pure quantum model is not a good one to account for typical probabilistic data exhibiting question order effects. Of course, one may object that by considering additional dimensions, and maybe degenerate measurements, it could still be possible to model the experimental data by means of the sole Born rule and the projection postulate. However, as rightly mentioned by \citet{Boyer-KassemEtal2015}: ``Introducing dimensions of degeneracy should be justified, so as not to be accused of being just ad hoc.'' But as we have shown, this is not necessary, as to exactly model the data one only needs to admit a more general class of measurements, also able to describe non-uniform fluctuations, as it can be naturally done in the GTR-model. The advantage and relevancy of using the latter is that it also allows handling another situation that is problematic for the standard quantum formalism: \emph{response replicability}. This is the topic we are going to address in the next sections.

\section{Response replicability}
\label{response replicability}

Quoting from \citet{KhrennikovEtal2014}: ``[...] quantum theory (QT) encounters difficulties in accounting for some very basic empirical properties. In opinion polling [...], there is a class of questions such that a repeated question is answered in the same way as the first time it was asked. This agrees with the L\"uders projection postulate [...]. In many situations, we also expect that for a certain class of questions the response to two replications of a given question remains the same even if we insert another question in between and have it answered. This property can only be handled by QT if the conventional observables representing different questions all pairwise commute, i.e., can be assigned the same set of eigenvectors. This, in turn, leads to a strong prediction: the joint probability of two responses to two successive questions does not depend on their order. This prediction is known to be violated for some pairs of questions. The explanation of the `question order effect' is in fact one of the most successful applications of QT in psychology \citep{WangBusemeyer2013}, but it requires non-commuting observables, and these, as we have seen, cannot account for the repeated answers to repeated questions.''

In the previous sections we have slightly ``resized'' this last statement regarding the success of standard quantum theory in accounting question order effects, as we have seen that the probability models these effects are able to generate are generally incompatible with the specific structure of the Born rule and can only be represented in more general quantum-like
frameworks, like the one provided by the GTR-model. As we are now going to explain, different from quantum theory, the GTR-model is also
able to jointly handle, in a consistent way, question order effects \emph{and} response replicability. The reason why this is possible is that in the GTR-model not only contexts can change states, through the mechanism of the breaking and collapsing of the elastic bands, but also states can change contexts, by allowing the probability distributions describing the way the elastics break to also depend on the outcomes obtained in the previous measurements,
exactly in the same way we humans, once we have formed an opinion, we don't need to form it again, by definition of what an opinion (and consequently an opinion poll) is.  

So, we start from the observation that there is a nonempty class of measurements such that, when they are performed in successive trials, if one of the measurements in the sequence is equal to a previous measurement, the same outcome that was previously obtained will be obtained again, but with certainty, i.e. with probability equal to $1$. For instance, considering the $A$ measurement consisting of the interrogative context ``Do you generally think Bill Clinton is honest and trustworthy?'' with outcome $Ay$ and $An$, a typical respondent, actualizing a non pre-existing answer the first time s/he is subjected to $A$, not having a predetermined opinion regarding the honesty and trustworthiness of Clinton, when subjected a second time to $A$, s/he will usually respond in the same way. In other terms, the possible outcomes of the sequential measurement $AA$ are only $AyAy$, and $AnAn$, as the outcomes $AyAn$ and $AnAy$ are generally not observed. By this we are not saying that all measurements are necessarily of this kind, but that, as remarked in \citet{KhrennikovEtal2014}, there certainly is a vast class of measurements for which this is the case. 

In quantum mechanics, measurements exhibiting this kind of immediate repeatability, i.e., not leading to a new outcome when they are repeated, are called (according to Pauli's classification) ``of the first kind,'' and are precisely those prescribed by the L\"uders-von Neumann projection postulate.\footnote{It is worth mentioning that in quantum physics many \emph{bona fide} measurements are not of the first kind, i.e., the post-measurement state will be in general substantially different from that predicted by the projection postulate. A simple example of a measurement that is not of the first kind is the counting of the photons' number of an electromagnetic field, by means of a photo-detector. Here the state after the measurement is always a vacuum state, irrespective of the measurement result. Another example is the measurement of the nuclear spin by free induction decay. In this case the state after the measurement is the thermal statistical mixture of upper and lower spin states, again irrespective of the measurement result. What is important to observe is that the validity of the Born rule does not depend on that of the projection postulate, and that the applicability of the latter needs to be justified by taking into account the specificities of the measurement protocol.} The more general class of measurements described in the GTR-model are also of the first kind, as is clear that when we perform the same measurement a second time, being the abstract point particle already located in one of the two end points of the elastic structure, its position cannot be further changed by the collapse of a new elastic, stretched along the same two end points. So, both the idealized measurements described in the standard quantum formalism, and those described in the GTR-model, exhibit this nice property of producing a stable change of the state of the entity, which can no longer change under the influence of the same measurement context (clearly an ideal situation to identify an outcome of an experiment by means of a so-called \emph{eigenstate}). 

The situation is different when the measurement $A$ is repeated not immediately after $A$, but after having performed an inter-measurement $B$, like the one corresponding to the question: ``Do you generally think Al Gore is honest and trustworthy?'' In other terms, we now consider the sequential measurement $ABA$, and according to our hypothesis, subjects will generally respond to the third $A$-question in exactly the same way they have responded to the first $A$-question, also when in between they have been subjected to the $B$-question. We are not interested here in explaining what are the psychological mechanisms behind this repeatability of the outcome (desire of coherence, learning, fear of being judged when changing opinion, etc.), but in understanding if it can be naturally modeled, in a way that is compatible with the observed question order effects. In other terms, compatibly with our previous modeling of the $AB$ and $BA$ measurements, we have to investigate if it is possible to model a measurement $ABA$, only producing as its
possible outcomes the four outcomes: $AyByAy$, $AyBnAy$, $AnByAn$ and $AnBnAn$, or equivalently a measurement $BAB$, only producing the four possible outcomes: $ByAyBy$, $ByAnBy$, $BnAyBn$ and $BnAnBn$.

As demonstrated by \citet{KhrennikovEtal2014}, this situation cannot be described by the standard quantum formalism, as replicability is only possible if the $A$ and $B$ measurements are described by commuting operators, which would then be in conflict with the existence of question order effects. Of course, as we have shown, the experimental incompatibility defined by non-commuting Hermitian operators is per se insufficient to correctly model the data, but certainly no order effects, of whatever kind, are possible for compatible observables. On the other hand, and different from quantum mechanics, the GTR-model allows for the description of `cognitive adjustments in accordance with previous experiences,' by naturally incorporating them in a process of change of the probability distributions describing the breaking of the elastic structures.

In that respect, let us come back to our hypothesis of weak compatibility, according to which the probability distribution describing the breaking of the $A$-elastic is the same if the measurement is performed as a first measurement, or following the $B$-measurement, and similarly that the probability distribution describing the breaking of the $B$-elastic is the same if the measurement is performed as a first measurement, or following the $A$-measurement. If we assume that the probability distribution $\rho_A$ describes the way subjects evaluate honesty and trustworthiness in relation to Clinton, then weak compatibility tells us that the process of answering the $B$-question does not alter their way of carrying out the evaluation (and same for $\rho_B$). This means that the observed order effects can be understood as resulting only from the ``change of perspective'' induced by answering $A$ before or after $B$, and not by a change in one's way of evaluating the situation. 

However, our weakly compatibility hypothesis does not say anything about a possible change of the probability distribution $\rho_A$ (resp. $\rho_B$) following the $AB$ (resp., $BA$) measurement. In case measurement $A$ is repeated a second time, in the direct sequence $AA$, we can assume for simplicity that $\rho_A$ doesn't change after the first measurement. In any case, the measurement being of the first kind, the point particle would not be able to further change its position, even if in the second measurement $\rho_A$ would have changed. But what about the change of $\rho_A$ following the $B$ measurement? In other terms, what should be the change induced by the sequential measurement $AB$, in order to  properly account for the replicability effect? 

Let $\rho_A(x)\equiv \rho_A(x|\psi)$ be the initial probability distribution describing the $A$-measurement. According to weak compatibility, we have $\rho_A(x|\psi)=\rho_A(x|\psi By)=\rho_A(x|\psi Bn)$, i.e., the probability distribution does not change if the $B$ measurement is performed first, whatever its outcome. However, here we are in the situation where it is the $A$-measurement that is performed first, in the sequence $AB$. So, to the four admissible
outcomes $AyBy$, $AyBn$, $AnBy$ and $AnBn$, we can let correspond to the following transitions for  both the states and the probability distributions:
\begin{equation}
\begin{array}{llll}
\psi\to Ay\to By, & \psi\to Ay\to Bn, & \psi\to An\to By, & \psi\to An\to Bn, \\
\rho_A\to\rho_A \to\rho_A^{AyBy}, & \rho_A\to\rho_A \to\rho_A^{AyBn}, & \rho_A\to\rho_A \to\rho_A^{AnBy}, & \rho_A\to\rho_A \to\rho_A^{AnBn},
\end{array}
\label{rhotransitions}
\end{equation}
where we have defined the truncated and renormalized probability distributions:
\begin{eqnarray}
&&\rho_A^{AyBy}(x) = {\rho_A(x)\over \int_{-1}^{\cos\theta} \rho_A(x)dx} \chi_{[-1,\cos\theta)}(x),\quad \rho_A^{AyBn}(x)={\rho_A(x)\over \int_{-1}^{-\cos\theta} \rho_A(x)dx} \chi_{[-1,-\cos\theta)}(x),\nonumber\\
&&\rho_A^{AnBy}(x) = {\rho_A(x)\over \int_{\cos\theta}^{1} \rho_A(x)dx} \chi_{(\cos\theta,1]}(x),\quad \rho_A^{AnBn}(x)={\rho_A(x)\over \int_{-\cos\theta}^{1} \rho_A(x)dx} \chi_{(-\cos\theta,1]}(x),
\label{rhotransitions2}
\end{eqnarray}
with $\chi_I(x)$ the characteristic function of the interval $I$. Consider for instance the probability $P(AyByAy|\psi)$, for the outcome $AyByAy$, in the sequential measurement $ABA$. It is clearly given by the product of the three conditional probabilities: $P(Ay|\psi AyBy) P(By|\psi Ay) P(Ay|\psi)$, with: 
\begin{equation}
P(Ay|\psi AyBy)=\int_{-1}^{\cos\theta}\rho_A^{AyBy}(x)dx = {\int_{-1}^{\cos\theta}\rho_A(x)dx\over \int_{-1}^{\cos\theta} \rho_A(x)dx} =1,
\end{equation}
in accordance with our hypothesis of replicability, and same thing for the other admissible outcomes. 

\begin{figure}[!ht]
\centering
\includegraphics[scale =.4]{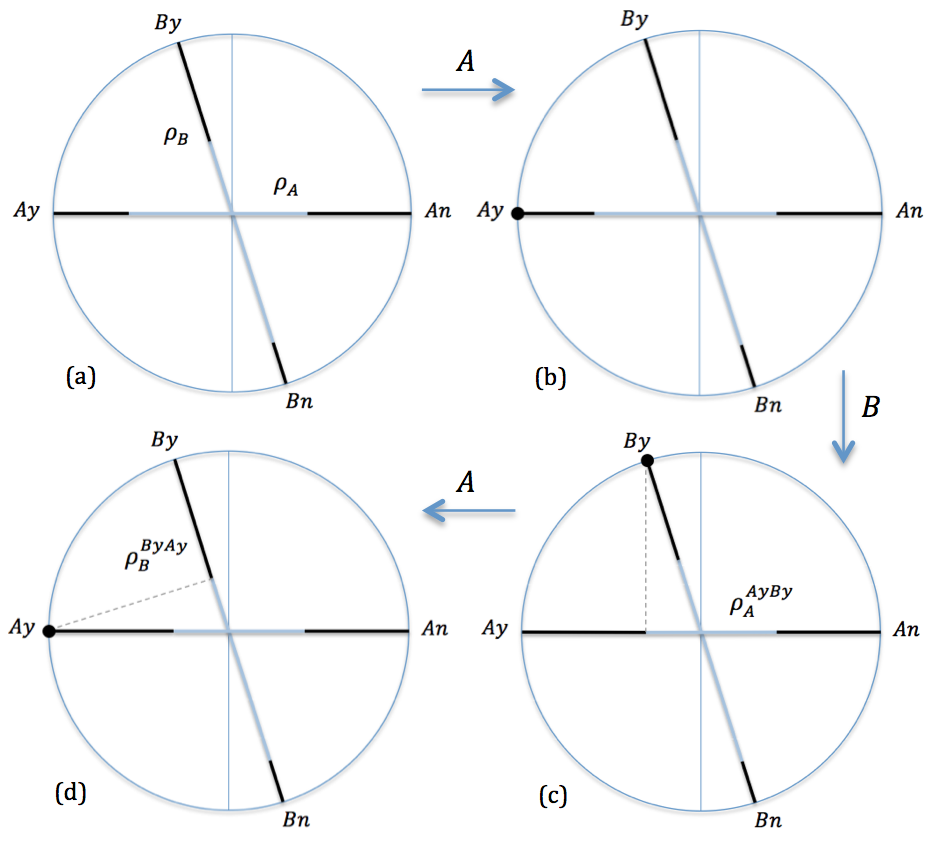}
\caption{The sequential measurement $ABA$. Figure (a) represents the situation before the first $A$-measurement. The representative point particle is located in ${\bf x}_\psi$, at the surface of the sphere, outside of the plane of the two elastic bands describing the measurements $A$ and $B$ (and therefore cannot be represented in the figure), here described by the probability distributions $\rho_A$ and $\rho_B$, given by (\ref{locallyuniform-rho}), with the parameters taking the values (\ref{parameters-values-Clinton}) of the Clinton/Gore data. Following the first $A$-measurement, we have assumed that the indeterministic breaking of the elastic has produced outcome $Ay$, as indicated in Figure (b). Then, in the second $B$-measurement, we have assumed that the transition $Ay\to By$ has occurred, to which, according to (\ref{rhotransitions})-(\ref{rhotransitions2}), also corresponds the transition: $\rho_A\to \rho_A^{AyBy}$, as illustrated in Figure (c). The outcome $Ay$ of the third $A$-measurement, described in Figure (d), is now certain in advance, in accordance with the hypothesis of replicability, and this $By\to Ay$ deterministic transition is associated with the probability distribution change: $\rho_B\to \rho_B^{ByAy}$. Then, subsequent $B$-measurements and $A$-measurements will not change anymore the nature of the elastic bands and state transitions will be perfectly deterministic.
\label{ABA}}
\end{figure}

Figure~ \ref{ABA} illustrates this process. Of course, the same holds true, \emph{mutatis mutandis} for the changes of the probability distribution $\rho_B$ in the sequence of measurements $BAB$. This means that once we have performed either $AB$ or $BA$, the outcomes of additional $A$ or $B$ measurements will become perfectly deterministic. For instance, if the measurement $A$ gave $Ay$, and the subsequent measurement of $B$ gave $Bn$, then, when performing again the $A$-measurement, the associated elastic band will be characterized by $\rho_A^{AyBn}(x)$, and if we perform again the $B$-measurement, the associated elastic band will be characterized by $\rho_B^{BnAy}(x)$ (with obvious notation), and from this point forward the probability distributions describing the two measurements will not change any more, so that subsequent alternations of the measurements $A$ and $B$ will only cause the point particle to transition from (the unit vector representative of) $Ay$ to $Bn$, to $Ay$, to $Bn$, and so on, in a perfectly deterministic way, in accordance with our expectation that answers will generally be repeated once they have been actualized.

It is worth observing that the prescription (\ref{rhotransitions})-(\ref{rhotransitions2}), about how the probability distribution $\rho_A$ has to change, in order to assure replicability of the obtained answer when $A$ is repeated, following a $B$-measurement, is a minimalistic one (no more than necessary), and is somehow reminiscent of the prescription of the L\"uders-von Neumann projection postulate for the quantum states transitions. Indeed, we just make unbreakable that specific segment of the $A$-elastic band that would produce the unwanted transition, but we do not alter the way the elastic can break everywhere else. This means that we also leave open the possibility that a new measurement $C$, once performed, say following the sequence $AB$, could again make a subsequent $A$-measurement indeterminate. In other terms, we consider the possibility that a measurements $C$ exist such that when $A$ is performed again, following the sequence $ABC$, the outcome could be different than what was obtained in the first $A$-measurement. 

As an example, imagine that $C$ corresponds to the following question: ``Newspapers report today a new scandal regarding Bill Clinton: does this diminish your trust in him?'' Imagine that the answer to the first $A$-measurement was $Ay$, that the answer to the subsequent $B$-measurement was $By$, and that the answer to this additional $C$-measurement was also affirmative, i.e. $Cy$. Here of course it is not important if the statement that the newspapers reported a new scandal is true or not. What is important is that, at the pragmatic level, this way of formulating the question is able to confuse some of the respondents, so much so that one can expect that, when asking $A$ again, following the $Cy$ outcome, the two outcomes $Ay$ and $An$ to be again available to be actualized, with a certain probability, thus interrupting the replicability of the outcomes of the previous $A$-measurement. Also, one expects that if the answer to the $C$-question was $Cn$ (suggesting for instance that the respondent did not believe the alleged scandal reported, or supposedly reported, in the newspapers), the replicability of the $A$-measurements will be preserved. 

In the GTR-model a situation of this kind can still be described, by modeling the $C$-measurement by means of an elastic band oriented in such a way that when the point particle orthogonally ``falls'' onto the $A$-elastic from the position ${\bf c}_y$, describing outcome $Cy$, it will land inside the interval where $\rho_A^{AyBy}(x)$ is breakable, i.e., $\cos\theta'\equiv {\bf c}_y\cdot {\bf a}_y\in[\cos\theta, -\epsilon_A+d_A]$, so producing, again, an indeterministic process (the probabilities of which will depend on the exact value taken by the angle $\theta'$). In other terms, when the outcome of the $ABC$-measurement is $AyByCy$, we can consider the transition: $\rho_A^{AyBy} \to\rho_A^{AyByCy}$, where $\rho_A^{AyByCy}$ can simply be taken to be identical to $\rho_A^{AyBy}$ (no change), whereas when the outcome is $AyByCn$, we can consider the transition $\rho_A^{AyBy} \to\rho_A^{AyByCn}$, with: 
\begin{equation}
\rho_A^{AyByCn}(x) = {\rho_A^{AyBy}(x)\over \int_{-1}^{-\cos\theta'} \rho_A^{AyBy}(x)dx} \chi_{[-1,-\cos\theta')}(x),
\label{rhotransitions3}
\end{equation}
so that in this case an additional $A$-measurement will give $Ay$ with certainty. 

The above was just an example meant to illustrate the great versatility of the GTR-model in handling in a consistent way, compatibly with question order effects, situations not only of repeatability, but also of partial (interrupted) repeatability, when additional particular measurements are considered that are able to change the mindset of the respondents in a way that can break the previously obtained response replicability. 

On the other hand, if one is just interested in modeling idealized situations where once a measurement has been performed, say the $A$-measurement, the same answer will always be replicated, when the measurement is repeated, independently of the number and nature of the additional measurements that are performed before the repetition, then for the change of the associated $\rho_A$-probability distribution it is sufficient to consider the following simple rule: $\rho_A(x)\to\delta(x+1)$, in case the answer was $Ay$, and $\rho_A(x)\to\delta(x-1)$, in case the answer was $An$, where $\delta$ denotes the Dirac delta function. In other terms, following the $A$-measurement, with initial probability distribution $\rho_A$, a subsequent $A$-measurement will simply
be described by an elastic band that is only breakable in its end point opposite to the previously obtained outcome, thus guaranteeing that it will be replicated for all pre-measurement states.

\section{Interpreting the model}
\label{interpretational framework}

In the previous sections we have shown that question order effects and response replicability can both be compatibly described in the GTR-model. This is because, different from the standard quantum formalism, the GTR-model provides an explicit representation of the measurements, by means of a mechanism of actualization of potential measurement-interactions. This mechanism opens to the possibility of associating different processes of actualization of a potential outcomes to a same set of outcomes, as described by the different possible $\rho$-probability distributions associated with an elastic band with a given orientation in the Bloch sphere, whilst in the quantum formalism only the uniform probability distribution is admitted. 

In other terms, the GTR-model, thanks to its greater structural richness, allows for a distinction between `a question and its possible answers,' and `the way a respondent selects one of these answers.' Different respondents are in principle associated with different set of probabilities for the outcomes, characterizing their `personal way of choosing an answer,' whereas in the quantum formalism the different subjects all necessarily choose in the same way, as if they were perfect ``Bornian clones''. This means that when we use the quantum formalism, order effects can only be described by means of the relative orientations of the different states in the Bloch sphere. This however constitutes a severe shortcoming in the description of psychological measurements (and in fact, possibly also in the description of physical measurements, but this is another story), as the Hilbertian geometry is too specific to account for the experimentally observed probabilities. 

As we have explained in \cite{AertsSassolideBianchi2015a, AertsSassolideBianchi2015b}, the breaking of the elastics (and more generally the disintegration of the hyper-membranes, when more general $N$-outcome interrogative contexts are considered) expresses in an intuitive way what we humans can typically feel when confronted with decisional contexts. Indeed, when a subject is confronted with a question (and more generally with a decision), say the $A$-question ``Do you generally think Bill Clinton is honest and trustworthy?'' and the associated possible answers ``Clinton is honest and trustworthy'' ($Ay$) or ``Clinton is not honest and trustworthy'' ($An$), this will automatically build a mental (neural) state of equilibrium, which results from the balancing of the tensions between the initial state (we will make precise in a moment what this initial state is about) and the two mutually excluding and competing answer-states. 

The creation of this mental state of equilibrium is described by the abstract point particle plunging into the sphere and reaching its on-elastic position, giving rise to two ``tension lines'' going from its position to the two end points, representative of the two competing outcomes. At some moment, this mental equilibrium will be disturbed, in a non-predictable way, and the disturbance will cause an irreversible process during which, almost instantaneously, the particle will be drawn to one of the two vertices of the elastic (representing the answers). This ``symmetry breaking'' process is described in the model by the random manifestation of a breaking point, which by causing the collapse of the elastic band breaks the tensional equilibrium that had previously been built, hence the name ``tension-reduction'' given to the model. 

If the elastic band is representative of an aspect of the mind of the subject, here generally understood as a memory structure sensitive to meaning, how should we interpret the abstract point particle, which interacts with it? In other terms, how should we interpret the initial state and the final
outcome states? From the perspective of the GTR-model, they should not be interpreted as a `description of the subject's beliefs,' on the issue in place, but
as a `description of an element of conceptual reality which is independent of the human mind of the person that can possibly interact with it.' In the case of the Clinton/Gore experiment, this element of conceptual reality is nothing but the conceptual entity `honesty and trustworthiness,' which for simplicity, in the following, we shall simply denote `honesty.' 

Prior to the measurement, this conceptual entity is in its ``ground'' (most neutral) state $\psi$, and this will be true for all the respondents, as prior to the measurement no specific contextualization of `honesty' is given. Then, when performing measurement $A$, subjects are asked if they think Clinton is honest. This question can be reformulated/reinterpreted in the following equivalent way: ``What best represents `honesty,' between the two possibilities: `Clinton is honest' ($Ay$) and `Clinton is not honest' ($An$).'' Now, `Clinton is honest' and `Clinton is not honest' are again to be considered as descriptions of elements of conceptual reality that are independent of the subjects' minds, and since they correspond to two different contextualizations of the conceptual entity `honesty,' they are to be interpreted as two different ``excited'' (non-neutral) states of said entity. 

So, we have distinguished three states of  the conceptual entity in question: the $\psi$-ground state `honesty,' the $Ay$-excited state `Clinton is honest', and the $An$-excited state `Clinton is not honest.' The reason why the state $\psi$ can be considered to be a superposition state, relative to the two (mutually exclusive states) $Ay$ and $An$, is not because, primarily, a human subject would be uncertain about the honesty or dishonesty of Clinton, but because `honesty' is available to be meaningfully contextualized as `Clinton is honest' or as `Clinton is not honest.' This is so because `honesty' is an abstract concept admitting a number of possible (less abstract) contextualizations, and the role of a human mind, when subjected to an interrogative context, is precisely that of actualizing one of these ``more concrete'' contexts, thus producing either the transition $\psi\to Ay$, or $\psi\to An$, generally in an unpredictable way. And this role of the human mind is made manifest in the GTR-model by means of the indeterministic dynamics of an abstract breakable elastic band. 

The nature of this breakable elastic band will generally depend not only on the two outcome-states that are available to be actualized, as the elastic will have these two outcome-states as its two end points, but also on the specific \emph{forma mentis} of the subject, described by its individual probability distribution $\rho_A^{(i)}$, where the superscript ``$(i)$'' identifies the $i$-th respondent among the $n$ respondents that are contributing to the overall statistics of outcomes.

Let us go one step further in our analysis by assuming that the outcome of the $A$-measurement was $Ay$, and that right after it was obtained, the $i$-th respondent is also subjected to the $B$-measurement. This corresponds to the question ``Do you generally think Al Gore is honest?'' As we have seen, this subsequent measurement is now performed not on the pre-measurement state $\psi$, but on the outcome-state $Ay$ of the previous measurement. To understand why it is correct to proceed in this way, we have to consider that since the $B$-measurement is performed immediately after the $A$-measurement, the answer $Ay$ will still be present in the `field of consciousness' of the respondent, and therefore will also play a role in the contextualization of the entity subjected to the interrogative process. In other terms, when the second $B$-measurement is performed, the conceptual entity is not anymore in the `honesty' ground state, but in the `Clinton is honest' excited state. So, the second $B$-measurement can be effectively rephrased as the following interrogation: ``What best represents `Clinton is honest,' between the two possibilities: `Gore is honest' ($By$) and `Gore is not honest' ($Bn$).''

To answer this second question, the respondent will again use a mental elastic band, which will be different from the previous one because the two outcome-states are of course different, but also because, possibly, the way these two outcome-states are chosen is different, i.e., the associated probability distribution $\rho_B^{(i)}$ is not necessarily equal to $\rho_A^{(i)}$. Assume then that the $B$-measurement has produced the $By$ outcome, and that right after it the $i$-th respondent is subjected again to the $A$-measurement. If the answers to the two previous questions are still present in the respondent's field of consciousness, we should consider that this third measurement is performed on the conceptual entity in the state ``Gore is honest \& Clinton is honest.'' This state, however, describes a condition that is stronger than that of the state ``Clinton is honest,'' and therefore cannot anymore be represented in a $3$-dimensional Bloch sphere, where all 2-outcome measurements are necessarily non-degenerate. 

The reason why in the GTR-model we can keep the description within the $3$-dimensional Bloch sphere, is that it is possible to transfer some information from the states to the measurements, by changing the measurements' probability distributions, as we have explained in Sec.~\ref{response replicability}. Intuitively, we can think of this process as a transfer from the conscious (manifest) to the subconscious (hidden) level, in accordance with the idea that very few cognitive elements can remain in the actual focus of a subject. This means that the outcome of the first $A$-measurement will exit the subject's field of consciousness, but will possibly be integrated in its memory structure, still exerting some influence in the form of a new, updated probability distribution. As a consequence, when the $i$-th respondent is subjected to the third $A$-measurement, following the $AB$ sequential measurement, the state of the conceptual entity will just be ``Clinton is honest,'' thus still describable by one of the two end points of the $B$-elastic, and as we have shown in the previous section the updated probability distribution for the $A$-measurement will then account for the expected respone replicability of the $Ay$ outcome of the first measurement.

\section{The averaging process}
\label{Averaging}

Having explained how the mathematical formalism of the GTR-model can naturally be interpreted, we must take the final step in our analysis and consider
the critical issue of the average over the different respondents. In a typical psychological measurement, data are collected from a certain number $n$ of participants, for instance about a thousand in the Clinton/Gore and Rose/Jackson experiments. As we mentioned in the Introduction, in a recent work
\citep{AertsSassolideBianchi2015a, AertsSassolideBianchi2015b} we have shown that if a sufficiently large number of respondents is considered, who explore a sufficiently wide spectrum of different `ways of choosing an answer,' then the average over all their answers will be well approximated by a pure quantum measurement. 

Our result, however, only holds for a single measurement context, and not necessarily for a sequential measurement context. In fact, considering the analysis of the previous sections, it is even to be expected that the equivalence between a universal average and a pure quantum measurement will generally not hold in a sequential situation, as suggested by the fact that the Clinton/Gore and Rose/Jackson paradigmatic data are very far from being well approximated by the Born rule (or better, the Born rule in combination with the projection postulate).

To clarify a bit further this issue, let us consider again the single measurement context of the Clinton's $A$-measurement. First of all, we have to observe that the fact that a universal average can explain the emergence of the Born rule does not mean that in practice the different respondents will necessarily choose their answers by exploring a wide spectrum of probability distributions $\rho_A^{(i)}$. Indeed, an effective quantum measurement can easily be obtained also by considering a collection of individuals all having a predetermined answer regarding Clinton's $A$-question. This means that even though each individual mind may respond in a deterministic way (using for instance elastics that can only break in one of their end points), their ``collective mind'' can nevertheless behave indeterministically, as a ``quantum machine.''

Of course, we are not saying that this is what happens in reality in the Clinton's $A$-measurement: we are just observing that there are a priori two different notions of indeterminism, one possibly manifesting at the level of the individual mind, and another possibly manifesting at the level of the ``collective mind,'' when the outcomes of the different individuals are considered together, without distinguishing them, by means of a uniform average. There are no doubts that $A$ is typically a question where the respondents, apart possible exceptions, will generally do not have an a predetermined opinion, so that their answers will literally be created during the very survey, in a way that will depend both on the specificities of their minds (as described by their $\rho_A^{(i)}$) and on the context (as described by the state of the conceptual entity, which according to our interpretation is the same for all the respondents, in the same way as the words written in the present article are the same for all its potential readers). 

Generally, we don't have access to the $\rho_A^{(i)}$ of the different subjects. Indeed, to retrieve part of an individual's $\rho_A^{(i)}$, we should be able to perform the $A$-measurement many times on a same person, considering different initial conditions, also finding a way to avoid response replicability (for instance by waiting enough time between two answers, so that the previous outcome has left the field of consciousness of the subject, and also, possibly, its memory). This risks to be an almost impossible measurement to realize. There is however an easier way to obtain some information about an individual's $\rho_A^{(i)}$: simply, by asking her/him to directly evaluate the outcome probabilities, for instance by answering questions like: ``between $1$ and $100$, what is the degree of truthfulness of the statement `Clinton is honest'?'' In other terms, an individual's mind is able not only to select outcomes, but also probabilities of outcomes.

We are not saying of course that the probabilities obtained in this way, once averaged, will necessarily produce the same probabilities than those obtained when the subjects are just asked to provide a single outcome, and the relative frequencies of the different outcomes are then calculated. But considering the strong meaning connection between the notion of ``truthfulness of a statement about honesty'' and that statement being ``a good representative of honesty,'' we should expect the final statistics to be quite similar. But our point here was just to emphasize that considering the nature of the question addressed to the respondents, when asked to estimate the probabilities they will generally give values different from $0$ and $1$, and this means that when they  actualize a single outcome, it is reasonable to expect that the process is genuinely indeterministic.

In fact, this conclusion also follows from the very observation of  the existence of question order effects. Indeed, if the respondents would know in advance the answers to the $A$ and $B$ questions, their order would be irrelevant to the answer (assuming here that the respondents do not change their minds during the survey). But since we know that the order is not irrelevant, we can deduce that the answers were actualized in a contextual way during the interrogative process, i.e., that they were created, and not just discovered. 

So, the assumption that each respondent uses a non trivial $\rho_A^{(i)}$ is a reasonable one. Given this, it is very important to distinguish the effective $\rho_A$, describing the working of the ``collective mind,'' and the  different $\rho_A^{(i)}$, $i=1,\dots,n$, describing the working of the individual minds participating in the survey. We have seen that for the Clinton/Gore and Rose/Jackson data the overall probability distribution $\rho_A$ has to be different from $\rho_B$, and also that $\rho_A$ and $\rho_B$ have to be both non-symmetric with respect to the origin of the Bloch sphere. The following question then arises: Does this mean that also for each participant $\rho_A^{(i)}$ has to be different from $\rho_B^{(i)}$, and that  they have to be non-symmetric?

In other terms: Is it possible that, after all, each participant does choose in the same way, when confronted with the $A$ and $B$ measurements, and also in a perfectly symmetrical way with respect to the two possible outcomes, and that the observed difference and asymmetry of the effective $\rho_A$ and $\rho_B$ distributions, describing the ``collective mind,'' is just a combined effect produced by the averaging of the outcomes and the sequentiality of the measurements? 

To tentatively answer this question, we consider a very simple situation, consisting in only two respondents ($n=2$). We assume that the individual probability distributions are locally uniform, symmetric, and that they are the same for the two measurements: $(\epsilon_A^{(i)}, d_A^{(i)})=(\epsilon_B^{(i)}, d_B^{(i)})\equiv (\epsilon_{i},0)$, $i=1,2$. According to (\ref{explicitsolutionAB}), we have for the probability of, say, outcome $AyBy$: 
\begin{equation}
P^{(i)}(AyBy|\psi)= {1\over 4}\left[1+(\cos\theta +\cos\theta_A){1\over \epsilon_i}+\cos\theta\cos\theta_A{1\over \epsilon_i^2}\right], \quad i=1,2.
\label{symmetricalprob}
\end{equation}
For the average probability $P(AyBy|\psi)\equiv{1\over 2}[P^{(1)}(AyBy|\psi)+P^{(2)}(AyBy|\psi)]$, we thus have: 
\begin{equation}
P(AyBy|\psi)= {1\over 4}\left[1+(\cos\theta +\cos\theta_A){\epsilon_1 +\epsilon_2\over 2\epsilon_1 \epsilon_2}+\cos\theta\cos\theta_A {\epsilon_1^2+\epsilon_2^2 \over 2\epsilon_1^2\epsilon_2^2}\right].
\label{average-n=2}
\end{equation}
Clearly, (\ref{average-n=2}) can only be written in the form (\ref{symmetricalprob}) if $(\epsilon_1 +\epsilon_2)^2=2(\epsilon_1^2+\epsilon_2^2)$, or equivalently, $(\epsilon_1-\epsilon_2)^2 =0$, i.e., if $\epsilon_1=\epsilon_2$. But since by hypothesis $\epsilon_1\neq\epsilon_2$, we have that the effective elastic band cannot be symmetric. For instance, if we choose: $\epsilon_1=1$, and $\epsilon_2=0.4$, $\cos\theta = 0.3$, $\cos\theta_A = 0.1$, and $\cos\theta_B = 0.2$, (\ref{explicitsolutionAB}) give the following probabilities for the first respondent (for simplicity, we don't write the ``$\psi$-conditional statement): $P^{(1)}(AyBy)= 0.3575$, $P^{(1)}(AyBn)=0.1925$, $P^{(1)}(AnBn)=0.2925$, $P^{(1)}(AnBy)=0.1575$,
$P^{(1)}(ByAy)= 0.39$, $P^{(1)}(ByAn)=0.21$, $P^{(1)}(BnAn)=0.26$, $P^{(1)}(BnAy)=0.14$. For the second respondent we have: $P^{(2)}(AyBy)= 0.546875$, $P^{(2)}(AyBn)=0.078125$, $P^{(2)}(AnBn)=0.328125$, $P^{(2)}(AnBy)=0.046875$, $P^{(2)}(ByAy) = 0.65625$, $P^{(2)}(ByAn)=0.09375$, $P^{(2)}(BnAn)=0.21875$, $P^{(2)}(BnAy)=0.03125$.

Calculating the average probabilites, we thus obtain: $P(AyBy)= 0.4521875$, $P(AyBn)=0.1353125$, $P(AnBn)=0.3103125$, $P(AnBy)=0.1021875$, $P(ByAy)= 0.523125$, $P(ByAn)=0.151875$, $P(BnAn)=0.239375$, $P(BnAy)=0.085625$. Inserting these valuese into (\ref{cos-d-B})-(\ref{difference-B}), choosing for instance the same value $\cos\theta = 0.3$, one obtains for the other parameters: 
$\epsilon_A\approx 0.59$, $\epsilon_B\approx 0.57$, $d_A\approx 0.02$, $d_B\approx 0.01$, $\cos\theta_B\approx 0.19$ and $\cos\theta_A\approx 0.13$. 

We thus see that already starting from two respondents, who are assumed to select their answers using in both measurements the same symmetrical and locally uniform elastic band, when we consider the average of their outcome probabilities, the resulting effective model requires the use of two different and non-symmetric elastic bands. We can also observe, but this was of course to be expected, that the abstract ``collective mind'' uses an effective description for the states that is different from that inter-subjectively employed by the individual minds, as is clear that the modeling of the averaged probabilities requires different, effective values for $\cos\theta_A$ and $\cos\theta_B$ ($0.13$ instead of $0.1$ and $0.19$ instead of $0.2$, respectively). 

Of course, the above considerations do not prove that at the individual level the different respondents actually use a single symmetric $\epsilon$-elastic band, when answering both the $A$ and $B$ questions. This we don't know. The only thing we know is that such hypothesis is not incompatible with the modeling of the experimental data that we have presented in this work. Regarding the issue of response replicability, discussed in Sec.~\ref{response replicability}, it is worth observing that the effective ``collective mind'' will behave according to response replicability only if each one of the different individual minds will do so. This is so because if, say, $P^{(i)}(AyByAy) =P^{(i)}(AyBy) $, for all $i=1,\dots,n$, then this will also be trivially the case for the average: $P(AyByAy) ={1\over n}\sum_{i=1}^n P^{(i)}(AyByAy) ={1\over n}\sum_{i=1}^n P^{(i)}(AyBy) = P(AyBy)$. In other terms, response replicability is straightforwardly transferred form the individual to the collective level.

\section{Concluding remarks}
\label{concluding remarks}

It is time to summarize our findings. We have considered some paradigmatic data exhibiting question order effects and we have shown that they are genuinely non-Hilbertian, i.e., that they require a more general framework than that of quantum mechanics, at least if one wants the modeling to remain within the ambit of $2$-outcome, non-degenerate measurements. This more general framework is provided by the GTR-model, where
different from quantum mechanics, various typologies of measurements, associated with a same set of outcome-states, can be described. 

In order to exactly solve the GTR-model, we have considered three simplifying hypothesis: weak compatibility, local uniformity, and sensitivity to pre-measurement state. The solution obeying these three hypotheses remains sufficiently general to allow for an exact fitting of the experimental data. When this is done, one finds that the probability distributions describing the two sequential measurements not only need to be different from one another, but also asymmetric. This, however, could very well be a byproduct of the averaging over the different respondents. Indeed, even if we assume that each respondent selects in a non-contextual way the outcomes (using the same probability distribution in the two sequential measurements), without favoring one of the outcomes (using a parity invariant probability distribution), to the extent that different respondents are characterized by different probability distributions, contextuality and asymmetry will necessarily emerge at the level of the more abstract ``collective mind.'' 

We have also discussed replicability of the outcomes, showing that it can be naturally accounted in the GTR-formalism by allowing the measurements' probability distributions to change at the individual level (and consequently also at the collective level), when measurements are repeated, and and that this process of change of the probability distributions, which adds to the process of change of the states, remains compatible with the observed question order effects. 

We have also proposed the adoption of a non-subjectivist interpretation of the GTR-model formalism, and a fortiori of the quantum formalism, where states do not describe ``states of mind,'' or ``states of beliefs,'' but the objective contextual condition of the conceptual entity that is
subjected to the measurement, whose reality is independent from the respondents' belief systems. A mind, instead, is that entity  interacting with the different conceptual entities, which are part of the human culture, by means of a mechanism of actualization of potential measurement interactions, as described in the formalism by the breaking of the elastic bands (or by the disintegration of higher-dimensional hyper-membranes, for more general measurement situations \citep{AertsSassolideBianchi2014, AertsSassolideBianchi2015a, AertsSassolideBianchi2015b, AertsSassolideBianchi2015c}). 

Being the explicit description of the measurements absent in the standard quantum formalism, this is of course the main reason why subjectivist interpretations exist, not only in quantum cognition, but also in physics (so-called QBism is a typical example \citep{FuchsEtal2014}). The GTR-model \citep{AertsSassolideBianchi2015a, AertsSassolideBianchi2015b}, and more specifically its Hilbertian extended Bloch representation \citep{AertsSassolideBianchi2014, AertsSassolideBianchi2015c}, provides a completed version of quantum mechanics, in which the explicit fluctuations originating in the interaction between the measured entity and the measuring system are explicitly represented, thus explaining the possible nature of quantum indeterminism (so providing a compelling solution to the measurement problem). This allows understanding the collapse of the state vector as an objective physical process, and consequently interpreting the state vector as a description of the reality of the physical entity under consideration, and not merely of the experimenter's beliefs about it. 

Similarly, if the probabilities emerging from psychological measurements are the result of objective fluctuations in the interaction between a human mind and a `meaning entity' prepared in a given context, as described in our beyond-quantum  GTR-model, then a state describes the real contextual condition of the conceptual entity, and not the beliefs of a human mind about it. Beliefs are stored in the memory structure of the judging human entity, and when they play a role they do so by altering the dynamics of the elastic bands describing the way answers are selected. But there are additional reasons for adopting a realistic interpretation of the mathematical formalism in quantum cognition. Different from physics, where the hidden-measurements remain an hypothesis awaiting experimental confirmation, in psychological measurements they are in part manifested by the very existence of the different participants in a survey, and their different way of responding. Last but not least, another important reason for adopting the proposed interpretational framework is that it has demonstrated its relevance in the modeling of concept combinations in a comprehensive theory of conceptual representation \citep{GaboraAerts2002, AertsGabora2005a, AertsGabora2005b}.

Let us conclude by briefly commenting on the possibility of describing question order effects (but not response replicability) by remaining within the strict confines of standard quantum mechanics, by increasing the number of dimensions. For this, we observe that generally the order of different statements contained in a sentence is relevant for what concerns its perceived meaning. Take the following example \citep{Rampin2005}: 

\begin{quote}
A novice asked the prior:\\ 
``Father, can I smoke when I pray?'' \\
And he was severely reprimanded.\\
A second novice asked the prior:\\ 
``Father, can I pray when I smoke?''\\
And he was praised for his devotion.
\end{quote}

We see that ``pray \& smoke'' does not elicit the same potential meanings as ``smoke \& pray.'' In the same way, the potential meanings of the sentence: ``Clinton is honest \& Gore is honest,'' are not the same as for the sentence: ``Gore is honest \& Clinton is honest.'' The difference  is this time subtler, and not everybody will consciously perceive it, but, nevertheless, it is an objective difference. This means that when we perform the sequential measurement $F=AB$, and obtain the four outcomes: $AyBy$, $AyBn$, $AnBy$ and $AnBn$, these outcomes will correspond to states that are \emph{all different} from those associated with $G=BA$, i.e.: $ByAy$, $ByAn$, $BnAy$ and $BnAn$. 

From a quantum mechanical point of view, this means that $F$ and $G$ are necessarily described by two different non-commuting Hermitian operators, acting on $\compl^4$, and although we know that, in practice, $F$ and $G$ are executed as two-step processes, i.e., as processes during which an outcome state is created in a sequential way, we are not allowed to experimentally disentangle such sequence into two separate measurements, in what is called a product measurement. This in particular because product measurements are formed by commuting sub-measurements, and commutation will generally prevent the modeling of order effects. Therefore, the only way to go, to possibly obtain a pure quantum modeling of question order effects, is to resort to the notion of `entangled measurements,' as it was considered for instance in \citet{AertsSozzo2014a, AertsSozzo2014b}. We plan to come back to this possible alternative modeling strategy in future works.

\end{document}